# SteuerLLM: Local specialized large language model for German tax law analysis

Sebastian Wind (1,2,3), Jeta Sopa (1), Laurin Schmid (1,4), Quirin Jackl (5), Sebastian Kiefer (3), Fei Wu (1), Martin Mayr (1,2), Harald Köstler (2,6), Gerhard Wellein (2), Andreas Maier (1,2), Soroosh Tayebi Arasteh (1,7,8)

(1) Pattern Recognition Lab, Friedrich-Alexander-Universität Erlangen-Nürnberg, Erlangen, Germany.
(2) Erlangen National High Performance Computing Center, Friedrich-Alexander-Universität Erlangen-Nürnberg, Erlangen, Germany.
(3) DATEV eG, Nuremberg, Germany.
(4) Bavarian AI Taxation Laboratory, Department of Computer Science, University of Technology Nuremberg, Nuremberg, Germany
(5) Chair for Tax Law and Public Law, Friedrich-Alexander-Universität Erlangen-Nürnberg, Nuremberg, Germany.
(6) Chair of Computer Science 10, Friedrich-Alexander-Universität Erlangen-Nürnberg, Erlangen, Germany.
(7) Lab for AI in Medicine, RWTH Aachen University, Aachen, Germany.
(8) Department of Diagnostic and Interventional Radiology, University Hospital RWTH Aachen, Aachen, Germany.

**Correspondence**

Sebastian Wind, MSc (sebastian.wind@fau.de) or
Soroosh Tayebi Arasteh, Dr.-Ing. Dr. rer. medic. (soroosh.arasteh@rwth-aachen.de)
Pattern Recognition Lab
Friedrich-Alexander-Universität Erlangen-Nürnberg
Martensstr. 3
91058 Erlangen, Germany

## Abstract

Large language models (LLMs) demonstrate strong general reasoning and language understanding, yet their performance degrades in domains governed by strict formal rules, precise terminology, and legally binding structure. Tax law exemplifies these challenges, as correct answers require exact statutory citation, structured legal argumentation, and numerical accuracy under rigid grading schemes. We algorithmically generate SteuerEx, the first open benchmark derived from authentic German university tax law examinations. SteuerEx comprises 115 expert-validated examination questions spanning six core tax law domains and multiple academic levels, and employs a statement-level, partial-credit evaluation framework that closely mirrors real examination practice. We further present SteuerLLM, a domain-adapted LLM for German tax law trained on a large-scale synthetic dataset generated from authentic examination material using a controlled retrieval-augmented pipeline. SteuerLLM (28B parameters) consistently outperforms general-purpose instruction-tuned models of comparable size and, in several cases, substantially larger systems, demonstrating that domain-specific data and architectural adaptation are more decisive than parameter scale for performance on realistic legal reasoning tasks. All benchmark data, training datasets, model weights, and evaluation code are released openly to support reproducible research in domain-specific legal artificial intelligence. A web-based demo of SteuerLLM is available at [https://steuerllm.i5.ai.fau.de](https://steuerllm.i5.ai.fau.de).

# Introduction

Large language models (LLMs) have demonstrated strong capabilities across a wide range of language understanding and reasoning tasks[1–4]. However, their performance remains fragile in domains governed by strict formal rules, precise terminology, and legally binding structure[5,6]. Law, and tax law in particular, exemplifies these constraints. Correct tax law reasoning requires exact statutory citation, hierarchical interpretation of interdependent legal norms, structured written argumentation, and numerical accuracy under rigid rules. In this setting, seemingly minor errors can invalidate an otherwise plausible answer[7]. As a result, general-purpose instruction-tuned[8] models often fail to meet the academic and professional standards required in tax law[5,9].

Prior work in legal artificial intelligence (AI) has explored retrieval-augmented generation[10,11], prompt engineering[4,5], or fine-tuning on curated legal corpora[12–14]. While these approaches can improve factual recall and stylistic alignment, they are typically evaluated on synthetic datasets or narrowly scoped tasks that do not reflect real assessment conditions[11,15]. Many existing benchmarks[11,16–18] emphasize short answers, classification, or binary correctness and therefore fail to capture the graded, partial-credit structure of legal examinations, where correctness is incremental and tightly coupled to statutory precision. This provides limited insight into whether language models can perform robust legal reasoning under realistic, high-stakes constraints. German tax law provides a particularly demanding test case[19,20]. It is highly codified, frequently amended, and characterized by dense cross-references between statutory provisions and detailed numerical rules. University tax law examinations are explicitly designed to reflect these properties. They require students to integrate doctrinal knowledge, structured legal analysis, and numerical computation under strict grading schemes[21,22]. Performance on such examinations therefore offers a stringent and economically valid benchmark for legal reasoning in language models. Despite this, no open benchmark derived from authentic German tax law examinations has been available, and no domain-adapted model has been systematically evaluated across such material spanning multiple tax domains, semesters, and academic levels.

In this study, we address these limitations through two complementary contributions (**Figure 1**). First, we introduce SteuerEx, the first open benchmark constructed from authentic German university tax law examinations. SteuerEx consists of 115 expert-validated examination questions drawn from undergraduate- and graduate-level courses administered across multiple semesters between 2016 and 2024. The benchmark spans six core tax law domains, including corporate tax, income tax, value-added tax (VAT), fiscal procedure, partnership taxation, and foundational tax law. Reference solutions are decomposed into independently scorable legal statements with explicit point values, enabling fine-grained, partial-credit evaluation that mirrors real academic grading practice (**Figure 2**). This design captures not only final correctness, but also partial legal reasoning quality under authentic assessment conditions.

Second, we present SteuerLLM, a domain-adapted LLM for German tax law. SteuerLLM is trained using a large-scale synthetic dataset generated from authentic examination material through a controlled retrieval-augmented pipeline grounded in statutory texts and authoritative legal sources. To incorporate domain-specific capacity without degrading general language and reasoning abilities, we employ a block expansion strategy that adds trainable Transformer layers while largely preserving pretrained parameters. This architectural choice allows specialization through additional depth rather than full fine-tuning. We evaluate SteuerLLM alongside a broad spectrum of instruction-tuned and reasoning-oriented LLMs, ranging from 3B to over 600B parameters, including both open-weight and proprietary systems (**Table 1**). Across the SteuerEx benchmark, results show that domain-specific training and architectural adaptation are more decisive for performance on authentic tax law examinations than parameter scale alone. A 28B-parameter SteuerLLM consistently outperforms substantially larger general-purpose models, while smaller domain-adapted variants remain competitive with mid-sized baselines. Beyond aggregated scores, we analyze performance across tax law categories and compare model outcomes with anonymized student examination results aggregated by domain. This comparison provides a concrete reference point for interpreting model competence relative to real academic performance, while highlighting persistent gaps in complex, high-stakes areas.

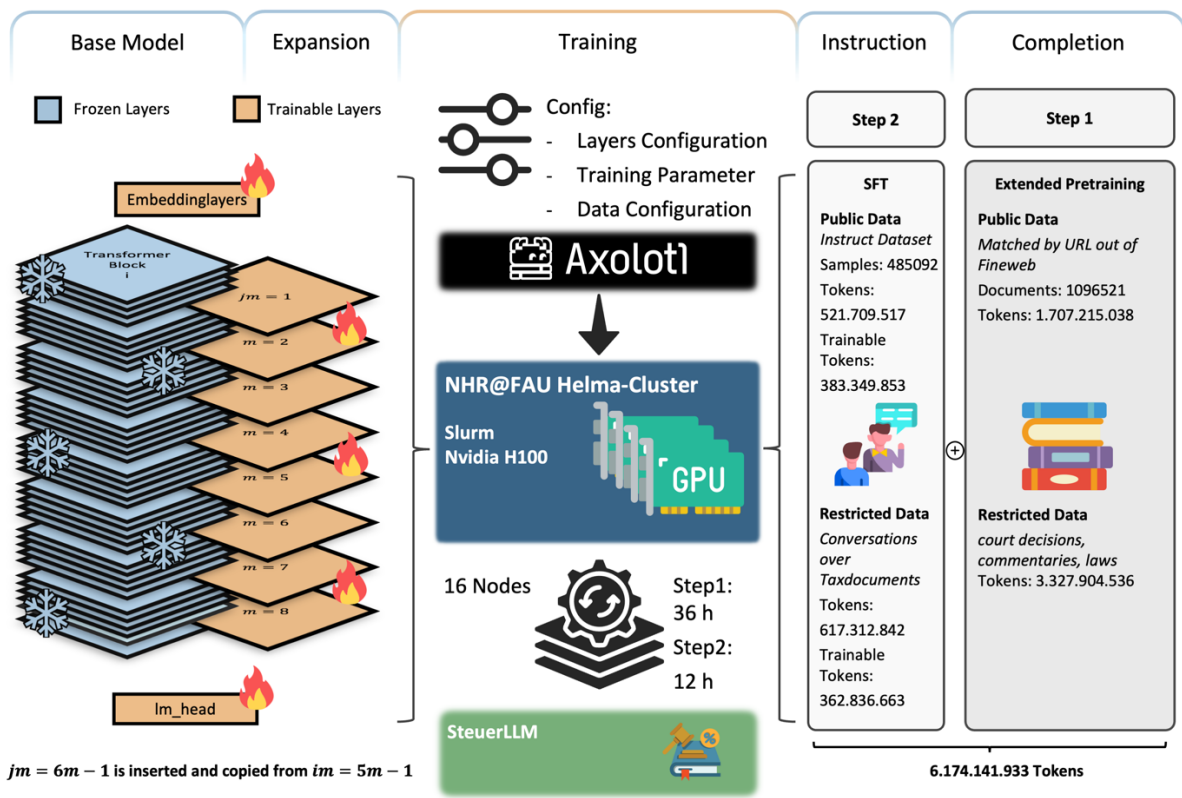

**Figure 1**: Architecture and training pipeline of SteuerLLM. The figure illustrates the model extension and training workflow used to construct SteuerLLM, which contains 28 B parameters. Starting from a pretrained transformer base model, additional embedding layers are inserted at regular depth intervals using a block extension strategy, while the original parameters remain frozen. The expanded model is trained using the Axolotl framework on a GPU cluster, combining large-scale public instruction data with restricted, domain-specific tax law data. Training proceeds in staged instruction and completion phases, integrating both conversational supervision and document-based learning. This design enables the incorporation of tax-specific knowledge while preserving general reasoning capabilities, resulting in the final 28B-parameter SteuerLLM model.

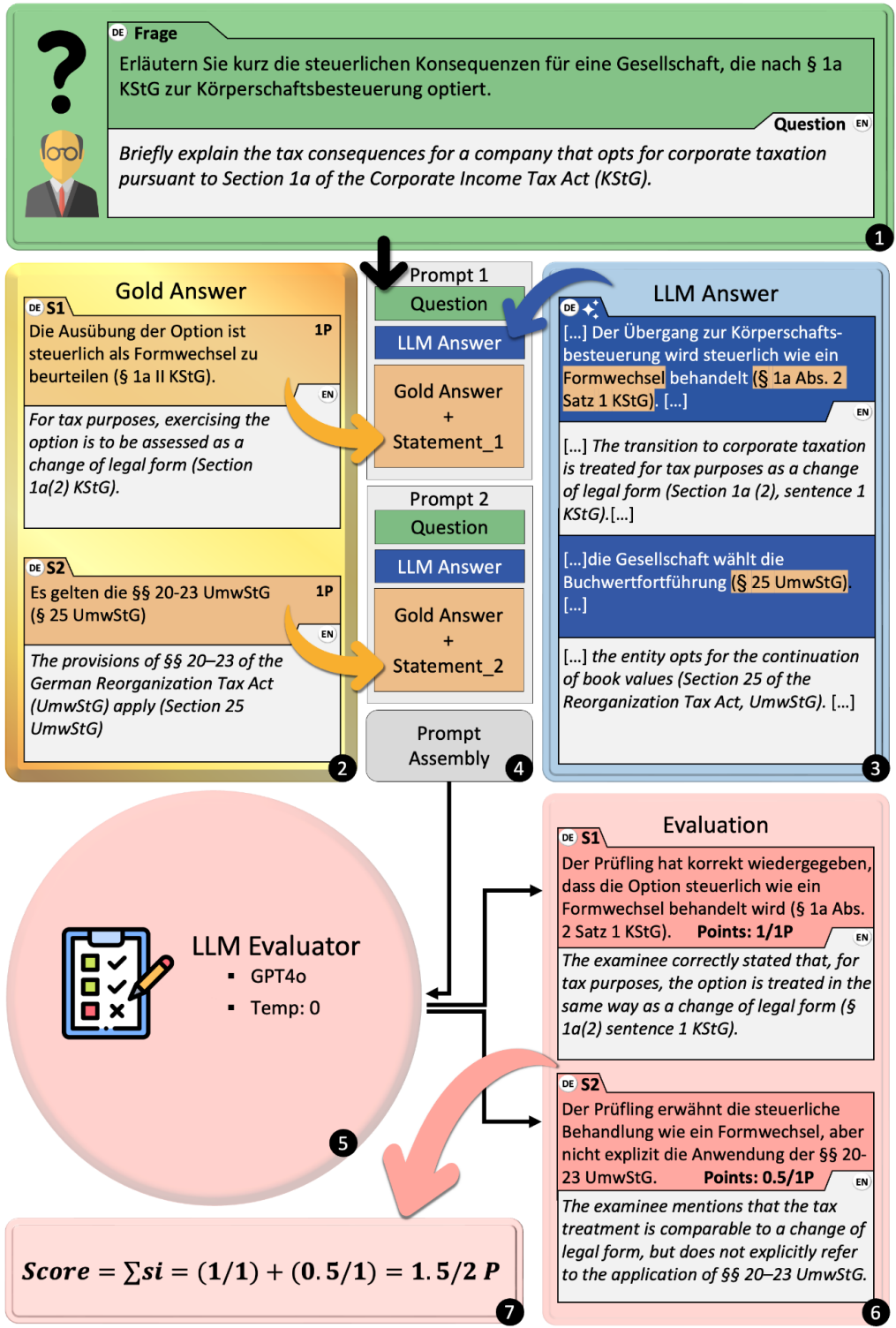

**Figure 2**: SteuerEx answering and evaluation workflow. Overview of the end-to-end evaluation procedure used in the SteuerEx benchmark. An exam-style tax law question is presented to the model (1). The expert reference solution is decomposed into discrete graded legal statements with assigned point values (2). The model generates a free-form answer (3), which is iteratively paired with each reference statement through a structured prompt assembly (4). A secondary LLM acting as evaluator assesses conceptual correctness, statutory accuracy, and completeness for each statement (5–6), awarding full or partial credit where appropriate. The final exam score is computed as the sum of statement-level scores, closely mirroring real university grading practices and enabling fine-grained assessment of partially correct legal reasoning (7).

All components of this study, including the SteuerEx benchmark, synthetic training data, model weights, and evaluation code, are released openly. By grounding evaluation in authentic examinations, analyzing a diverse set of modern LLMs, and employing transparent, reproducible methodology, this work establishes a rigorous framework for studying domain-specific legal reasoning in language models. More broadly, it demonstrates how realistic academic assessments can serve as high-fidelity benchmarks for evaluating generalization, specialization, and scaling behavior in deep learning systems.

**Table 1:** Specifications of the language models evaluated in this study. Summary of all LLMs assessed on the SteuerEx benchmark for German tax law reasoning. Listed for each model are the parameter count in billions, training category such as instruction-tuned (IT) or reasoning-oriented, accessibility, knowledge cutoff date, developer, and maximum context length in thousand tokens. The evaluated models span open-source, open-weights, and proprietary systems, and include both general-purpose instruction-tuned models and reasoning-focused architectures, as well as the proposed SteuerLLM variants. All locally deployed LLMs were assessed and used between January and April 2025, and the evaluations were performed from April 2025 until January 2026.

| Model name | Parameters (billions) | Category | Accessibility | Knowledge cutoff date | Developer | Context length (thousand tokens) |
|---|---|---|---|---|---|---|
| DeepSeek-R1-Distill-Llama-70B | 70 | Reasoning | Open-source | January 2025 | DeepSeek | 128 |
| Llama-3.2-3B-it | 3 | IT | Open-weights | December 2023 | Meta AI | 128 |
| Llama-3-8B-it | 8 | IT | Open-weights | March 2023 | Meta AI | 8 |
| Ministral-8B-it-2410 | 8 | IT | Open-source | October 2023 | Mistral AI | 128 |
| Mistral-Small-it-2409 | 24 | IT | Open-source | October 2023 | Mistral AI | 32 |
| Qwen2.5-14B-it | 14 | IT | Open-source | September 2024 | Alibaba Cloud | 131 |
| Qwen2.5-32B-it | 32 | IT | Open-source | September 2024 | Alibaba Cloud | 131 |
| Qwen2.5-3B-it | 3 | IT | Open-source | September 2024 | Alibaba Cloud | 32 |
| Qwen2.5-72B-it | 72 | IT | Open-source | September 2024 | Alibaba Cloud | 131 |
| Qwen2.5-7B-it | 7 | IT | Open-source | September 2024 | Alibaba Cloud | 131 |
| Small-SteuerLLM | 10 | IT | Closed-source | January 2025 | NHR@FAU | 32 |
| SteuerLLM | 28 | IT | Closed-source | January 2025 | NHR@FAU | 32 |
| Open-SteuerLLM | 28 | IT | Open-Source | January 2025 | NHR@FAU | 32 |
| DeepSeek-R1-671B | 671 | Reasoning, mixture of experts | Open-source | January 2025 | DeepSeek | 128 |
| Gemma-3-27B-it | 27 | IT | Open-weights | August 2024 | Google DeepMind | 128 |
| Gemma-3-4B-it | 4 | IT | Open-weights | August 2024 | Google DeepMind | 128 |
| GPT-4o-mini | Unknown | IT | Proprietary | October 2023 | Open-AI | 128 |

# Results

Before presenting the individual experimental results, we briefly summarize the newly-introduced evaluation benchmark and training data to provide context for the reported findings. SteuerEx is a benchmark designed to evaluate LLMs on realistic German tax law reasoning tasks using authentic university examinations. It consists of 115 examination questions with a total achievable score of 1,035.5 points, reflecting the weighted grading schemes used in real academic assessments. Each question is paired with an expert-validated reference solution that is decomposed into graded legal statements, allowing partial credit for incomplete but legally sound reasoning. The benchmark draws exclusively from undergraduate- and graduate-level tax law examinations administered at Friedrich-Alexander-Universität Erlangen-Nürnberg (FAU) from 2016 onward, with a focus on more recent exams to ensure alignment with the current legal framework. Questions span a wide range of German tax law domains and are structured to test not only factual knowledge, but also statutory interpretation, structured legal argumentation, and numerical accuracy, closely mirroring real examination and professional requirements.

## SteuerLLM outperforms general-purpose LLMs

We evaluated model performance on the SteuerEx benchmark using the normalized total exam score (percentage of 1,035.5 total points), with results summarized in **Table 2**. Overall, scores were substantially lower than what is typically observed on general-purpose natural language processing (NLP) benchmarks[23–26], highlighting that real-world German tax law examinations remain exteremly challenging for current instruction-tuned and reasoning-oriented LLMs. Most general-purpose open models scored below 25%, and several models in the 3B–14B range achieved only single-digit percentages, indicating that tax-law-specific statutory reasoning and structured answer requirements are not reliably captured by standard instruction tuning alone.

Across all evaluated systems (**Figure 3**), the strongest performance is achieved by DeepSeek-R1-671B, reaching 39% ± 3 (95% CI: 32–44; 399/1,035.5 points). Among the remaining models, SteuerLLM, with 28B parameters, attains 28% ± 2 (95% CI: 24–33), establishing the best result outside of DeepSeek-R1-671B. This performance exceeds all tested Qwen2.5 instruction-tuned baselines (all P < 0.0001), including substantially larger models such as Qwen2.5-72B-Instruct (19% ± 3) and Qwen2.5-32B-it (18% ± 2), and also outperforms GPT-4o-mini (22% ± 2, P = 0.0029) and Gemma-3-27B-it (23% ± 2, P = 0.0029). Notably, DeepSeek-R1-Distill-Llama-70B (20% ± 3) does not close the gap to SteuerLLM (P < 0.0001) despite having more than twice the parameter count, suggesting that distillation alone is insufficient to match a model trained with targeted domain data and tax-specific adaptation strategies. Small-SteuerLLM, despite being considerably smaller at 10B parameters, reaches 16% ± 2 (95% CI: 13–20) and performs competitively with several general-purpose models in the 14B–32B range, indicating that domain specialization yields measurable benefits even at reduced scale.

Differences relative to SteuerLLM are statistically significant for all other models where significance testing was performed (all P < 0.0001), supporting that the observed improvements are not explained by sampling noise. Overall, SteuerEx reveals a clear separation between general-purpose instruction-tuned LLMs and a tax-specialized model trained on targeted tax law data. While DeepSeek-R1-671B operates at approximately 24× the parameter count of SteuerLLM (671B vs. 28B) and remains the strongest overall system in this comparison (P = 0.0003 relative to SteuerLLM), SteuerLLM achieves the highest performance among all remaining evaluated models while operating at a markedly smaller scale than several baselines.

**Table 2:** Performance of language models on the SteuerEx benchmark. Scores are reported as normalized percentages of the maximum achievable score of 1,035.5 points and presented as mean ± standard deviation, with 95% confidence intervals shown in brackets. Absolute scores in points are reported alongside normalized values. Results are based on n=115 examination questions and estimated using bootstrapping with 10,000 repetitions and replacement while preserving pairing. P-values indicate statistical significance of each model's performance relative to SteuerLLM, computed using paired tests and adjusted for multiple comparisons where applicable. A p-value < 0.05 was considered statistically significant. N/A: not assigned.

| Model name | Score (normalized to percent) | Total points (out of 1035.5) | P-value (w.r.t. SteuerLLM) |
|---|---|---|---|
| DeepSeek-R1-Distill-Llama-70B | 20 ± 3 [15–25] | 204.0 | 0.0001 |
| Llama-3.2-3B-it | 9 ± 2 [6–13] | 90.0 | 0.0001 |
| Llama-3-8B-it | 8 ± 2 [4–11] | 79.0 | 0.0001 |
| Ministral-8B-it-2410 | 13 ± 2 [9–18] | 136.0 | 0.0001 |
| Mistral-Small-it-2409 | 20 ± 2 [17–25] | 212.0 | 0.0001 |
| Qwen2.5-14B-it | 12 ± 2 [9–16] | 129.0 | 0.0001 |
| Qwen2.5-32B-it | 18 ± 2 [14–22] | 186.0 | 0.0001 |
| Qwen2.5-3B-it | 5 ± 1 [3–7] | 54.0 | 0.0001 |
| Qwen2.5-72B-it | 19 ± 3 [14–24] | 196.0 | 0.0001 |
| Qwen2.5-7B-it | 12 ± 2 [9–14] | 120.0 | 0.0001 |
| Small-SteuerLLM | 16 ± 2 [13–20] | 171.0 | 0.0001 |
| SteuerLLM | 28 ± 2 [24–33] | 294.0 | N/A |
| Open-SteuerLLM | 23 ± 3 [18-29] | 241.0 | 0.1901 |
| DeepSeek-R1-671B | 39 ± 3 [32–44] | 399.0 | 0.0003 |
| Gemma-3-27B-it | 23 ± 2 [18–28] | 236.0 | 0.0029 |
| Gemma-3-4B-it | 11 ± 2 [8–14] | 113.0 | 0.0001 |
| GPT-4o-mini | 22 ± 2 [18–26] | 226.0 | 0.0029 |

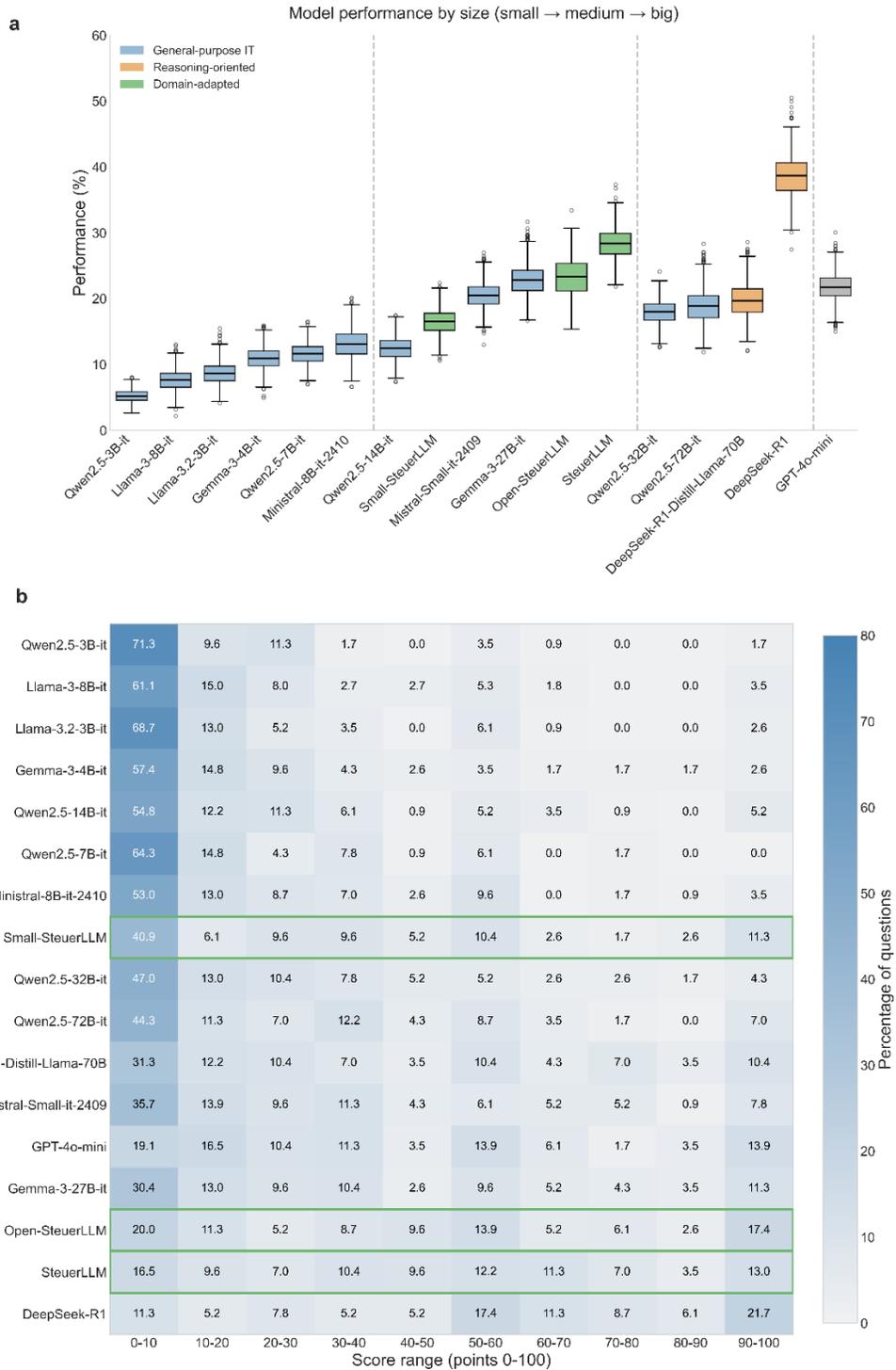

**Figure 3**: Model performance on the SteuerEx benchmark across scale and score distributions. **a** Normalized SteuerEx scores (percentage of the maximum achievable score) for all evaluated models, grouped by approximate parameter scale and ordered by median performance within each group. Boxplots summarize per-question scores, with boxes indicating the interquartile range, center lines the median, whiskers extending to 1.5×IQR, and points denoting outliers. Dashed vertical lines mark boundaries between size groups; model classes are color-coded. **b** Distribution of per-question normalized scores across predefined score bins (0–10 to 90–100) for each model, reporting the percentage of questions falling into each bin.

# Performance does not scale reliably with parameter count

To examine whether performance on SteuerEx scales with model size, we compared results across models spanning from 3B parameters to 671B parameters (**Table 2**). While larger models within the same family often outperform their smaller counterparts, parameter count alone does not reliably predict performance across model families. Several large instruction-tuned models, including Qwen2.5-72B-it (19% ± 3) and Qwen2.5-32B-it (18% ± 2), score significantly below SteuerLLM (28% ± 2) despite having up to 2.6× more parameters. These models also underperform Gemma-3-27B-it (23% ± 2), which is comparable in size to SteuerLLM. Conversely, mid-sized models such as GPT-4o-mini (22% ± 2) and Gemma-3-27B-it outperform multiple larger instruction-tuned baselines, further illustrating that scale alone is insufficient for strong tax law reasoning performance.

This pattern indicates that, on real-world German tax law examinations, increasing parameter count yields limited and inconsistent gains in the absence of domain-specific training or specialized reasoning objectives. Performance differences between models of similar scale frequently exceed differences attributable to size alone, suggesting that training data composition and optimization strategy are more decisive than raw parameter count in this setting. Even the largest model evaluated, DeepSeek-R1-671B achieves the highest overall score but does not establish a smooth scaling trend across the remaining models, highlighting the non-monotonic relationship between size and performance on SteuerEx.

Beyond mean scores, models also differ markedly in how performance is distributed across individual exam questions (**Figure 5**). Smaller instruction-tuned models are heavily concentrated in the lowest score bins, reflecting frequent near-zero or fragmentary answers. In contrast, SteuerLLM exhibits a broader score distribution, with a substantially higher proportion of responses achieving partial and mid-range credit, alongside occasional high-scoring answers. Large reasoning-oriented models such as DeepSeek-R1-671B show a further shift toward higher score ranges but retain non-trivial mass at low scores, indicating persistent difficulty on structurally complex cases. This distributional analysis suggests that domain-specific training primarily improves the likelihood of producing partially correct, exam-relevant legal reasoning rather than merely increasing the frequency of near-perfect answers.

# Tax law domain-specific performance patterns

To better understand where domain specialization helps most, we report category-level performance across all evaluated models (**Table 3**), using the six disjoint tax law categories that jointly constitute SteuerEx (**Table 4**). Performance varies substantially across domains, and differences between models are often larger within a category than what would be expected from parameter scale alone, indicating strong domain- and task-specific effects. Categories differ not only in size but also in internal doctrinal composition, with individual exams covering multiple statutory subtopics and point weightings (see **Supplementary Table 2** for a subtopic-level breakdown).

**Table 3:** Category-level performance of language models on the SteuerEx benchmark. Scores are reported as normalized percentages of the maximum achievable score within each tax law category and presented as mean ± standard deviation, with 95% confidence intervals shown in brackets. Maximum achievable points per category are: corporate tax (234.5 points), fiscal code (129 points), fundamentals of tax law (296 points), income tax (189 points), taxation of partnerships (66 points), and value-added tax (121 points). Results were estimated using non-parametric bootstrapping with 10,000 repetitions and replacement. The number of questions contributing to each category is: corporate tax (44), fiscal code (3), fundamentals of tax law (56), income tax (4), taxation of partnerships (4), and value-added tax (4). For categories with a small number of questions, bootstrap confidence intervals can be narrow or degenerate because resampling is constrained by the limited number of distinct observations, reducing the apparent variance despite genuine uncertainty.

| Model name | Corporate tax | Fiscal code | Fundamentals of tax law | Income tax | Taxation of partnerships | Value-added tax (VAT) |
|---|---|---|---|---|---|---|
| DeepSeek-R1-Distill-Llama-70B | 18 ± 3 [12, 25] | 3 ± 0 [3, 3] | 29 ± 4 [22, 36] | 22 ± 1 [22, 25] | 13 ± 5 [3, 17] | 26 ± 8 [21, 40] |
| Llama-3.2-3B-it | 6 ± 1 [4, 9] | 1 ± 0 [1, 1] | 12 ± 2 [8, 16] | 7 ± 1 [7, 11] | 4 ± 1 [2, 5] | 4 ± 2 [0, 5] |
| Llama-3-8B-it | 10 ± 3 [5, 15] | 0 ± 0 [0, 0] | 12 ± 2 [7, 16] | 8 ± 3 [7, 18] | 10 ± 3 [3, 12] | 30 ± 12 [23, 50] |
| Ministral-8B-it-2410 | 12 ± 3 [8, 17] | 0 ± 0 [0, 0] | 19 ± 3 [13, 25] | 19 ± 4 [18, 33] | 10 ± 7 [0, 15] | 19 ± 9 [14, 35] |
| Mistral-Small-it-2409 | 18 ± 3 [12, 24] | 10 ± 0 [10, 10] | 29 ± 3 [23, 35] | 26 ± 2 [25, 33] | 10 ± 7 [0, 14] | 27 ± 7 [23, 40] |
| Qwen2.5-14B-it | 12 ± 3 [7, 18] | 2 ± 0 [2, 2] | 17 ± 3 [11, 22] | 17 ± 2 [16, 22] | 15 ± 3 [6, 18] | 21 ± 8 [16, 35] |
| Qwen2.5-32B-it | 17 ± 3 [11, 23] | 5 ± 0 [5, 5] | 21 ± 3 [16, 27] | 22 ± 1 [21, 25] | 9 ± 3 [3, 11] | 31 ± 14 [23, 55] |
| Qwen2.5-3B-it | 7 ± 2 [4, 11] | 1 ± 0 [1, 1] | 8 ± 2 [5, 12] | 2 ± 1 [0, 3] | 3 ± 2 [0, 4] | 16 ± 5 [13, 25] |
| Qwen2.5-72B-it | 16 ± 3 [10, 22] | 7 ± 0 [7, 7] | 24 ± 3 [17, 31] | 31 ± 3 [30, 40] | 7 ± 2 [2, 8] | 31 ± 8 [27, 45] |
| Qwen2.5-7B-it | 9 ± 3 [4, 15] | 1 ± 0 [1, 1] | 15 ± 2 [10, 19] | 12 ± 1 [10, 13] | 8 ± 6 [0, 12] | 8 ± 1 [7, 10] |
| Small-SteuerLLM | 14 ± 3 [8, 20] | 10 ± 0 [10, 10] | 25 ± 4 [18, 32] | 13 ± 0 [13, 13] | 10 ± 7 [0, 15] | 33 ± 4 [31, 40] |
| SteuerLLM | 38 ± 4 [29, 46] | 14 ± 0 [14, 14] | 41 ± 3 [35, 47] | 24 ± 1 [24, 27] | 18 ± 8 [3, 23] | 35 ± 14 [27, 60] |
| Open-SteuerLLM | 38 ± 4 [30, 46] | 6 ± 0 [6, 6] | 42 ± 3 [36, 47] | 24 ± 1 [24, 29] | 11 ± 8 [0, 17] | 22 ± 5 [19, 30] |
| DeepSeek-R1-671B | 38 ± 4 [31, 46] | 20 ± 0 [20, 20] | 62 ± 3 [57, 66] | 32 ± 4 [31, 44] | 35 ± 15 [6, 57] | 40 ± 0 [40, 40] |
| Gemma-3-27B-it | 14 ± 3 [9, 20] | 9 ± 0 [9, 9] | 38 ± 3 [32, 44] | 22 ± 0 [22, 22] | 24 ± 6 [13, 28] | 37 ± 13 [30, 60] |
| Gemma-3-4B-it | 10 ± 2 [6, 15] | 4 ± 0 [4, 4] | 13 ± 2 [9, 18] | 8 ± 3 [7, 16] | 6 ± 4 [0, 9] | 32 ± 13 [25, 55] |
| GPT-4o-mini | 22 ± 3 [16, 29] | 11 ± 0 [11, 11] | 41 ± 3 [35, 48] | 23 ± 4 [22, 35] | 12 ± 9 [0, 18] | 26 ± 8 [22, 40] |

Across categories, DeepSeek-R1-671B attains the highest scores overall and is the only model that consistently reaches the top tier across all domains, with particularly strong performance in fundamentals of tax law (62% ± 3) and taxation of partnerships (35% ± 15). SteuerLLM achieves the strongest results among non-reasoning-specialized models across most categories and shows its largest advantage in domains that rely heavily on structured legal argumentation and multi-step statutory reasoning. In corporate tax, which is the largest category by question count (44 questions; **Table 4**), SteuerLLM reaches 38% ± 4 (95% CI: 29–46), matching DeepSeek-R1-671B (38% ± 4) and clearly exceeding all other instruction-tuned baselines, including Qwen2.5-72B-it (16% ± 3) and GPT-4o-mini (22% ± 3). Similarly, in fundamentals of tax law (56 questions), SteuerLLM reaches 41% ± 3 (95% CI: 35–47), outperforming most open instruction-tuned models and closely matching GPT-4o-mini (41% ± 3), although still below DeepSeek-R1-671B's substantially higher score. Performance is notably lower and more volatile in categories with few questions but high point density, where single questions can dominate outcomes. In fiscal code (3 questions), scores cluster in a narrow range for many models (e.g., several near 0–10%), while DeepSeek-R1-671B achieves 20% ± 0 and SteuerLLM reaches 14% ± 0.

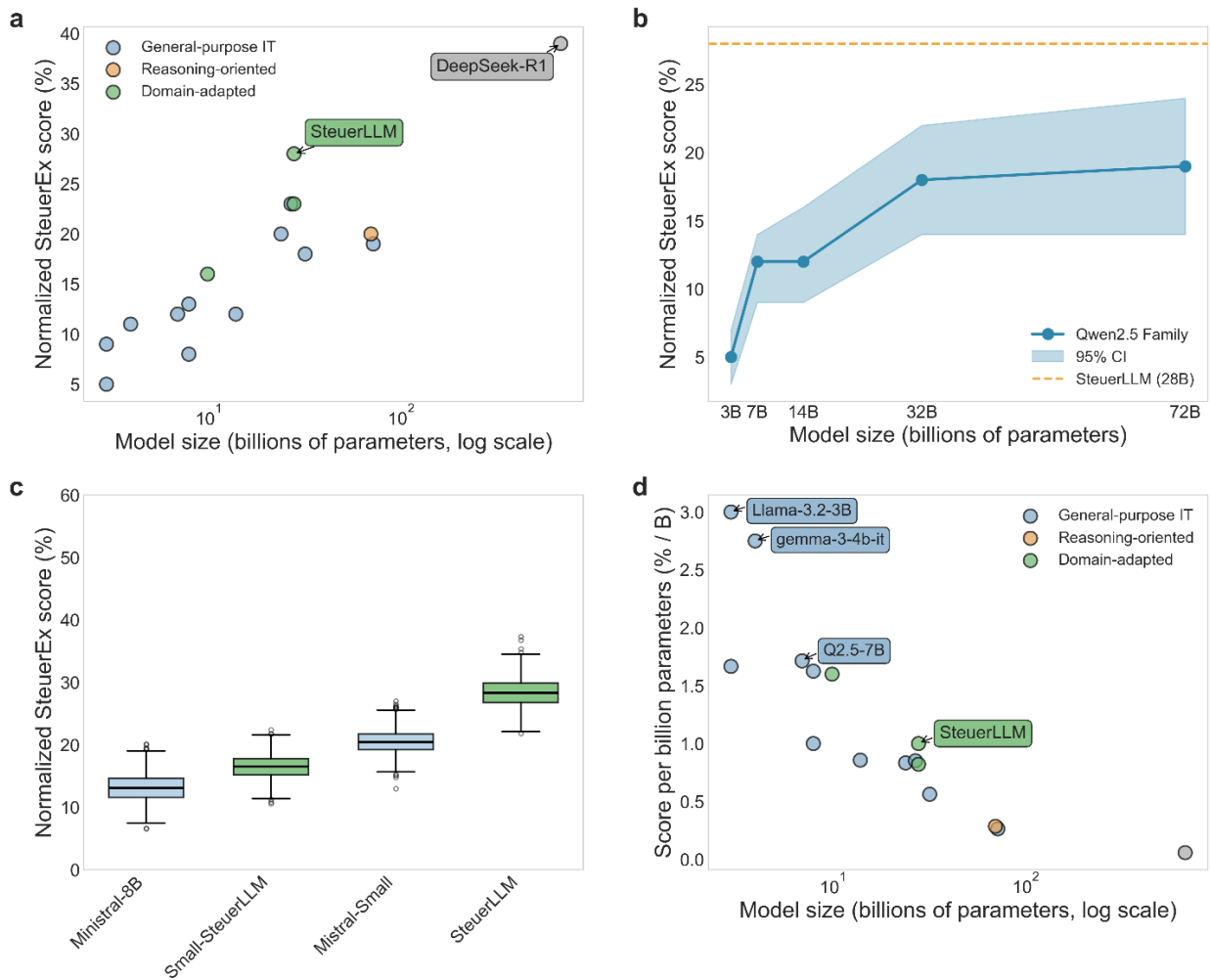

**Figure 4**: Scaling, distributional, and efficiency effects in German tax law reasoning. **a** Normalized SteuerEx score (percentage of the maximum achievable score) as a function of model size (billions of parameters, log scale) for all evaluated models, color-coded by model class. Selected models are annotated. **b** Within-family scaling behavior for the Qwen 2.5 instruction-tuned series, showing normalized score versus parameter count with shaded 95% bootstrap confidence intervals; the dashed line indicates SteuerLLM (28B). **c** Per-question normalized score distributions for representative models evaluated on identical question sets, shown as boxplots with boxes spanning the interquartile range, center lines indicating the median, and whiskers extending to 1.5×IQR. **d** Parameter efficiency measured as normalized SteuerEx score per billion parameters as a function of model size. All results are computed on the SteuerEx benchmark (n = 115 questions).

In income tax (4 questions), DeepSeek-R1-671B leads (32% ± 4), while multiple mid-to-large instruction-tuned models approach the mid-20% range, including Qwen2.5-72B-it (31% ± 3), Mistral-Small-it (26% ± 2), and SteuerLLM (24% ± 1). In taxation of partnerships (4 questions), DeepSeek-R1-671B again shows the strongest results (35% ± 15), followed by Gemma-3-27B-it (24% ± 6) and SteuerLLM (18% ± 8), highlighting the difficulty of highly specialized partnership cases even for domain-adapted models. Finally, value-added tax (VAT) exhibits the highest scores among instruction-tuned baselines, where SteuerLLM reaches 35% ± 14, Small-SteuerLLM attains 33% ± 4, and DeepSeek-R1-671B achieves 40% ± 0. The results reinforce that SteuerLLM's gains are not uniform across tax law, but concentrate in high-coverage domains that emphasize structured reasoning and statutory grounding, while narrow categories with few questions remain challenging and yield unstable estimates due to limited sampling support (**Table 4**).

## Comparison with student examination performance across tax law domains

To contextualize SteuerLLM performance relative to human examinees, we compared category-level model scores against anonymized student outcomes from FAU tax law examinations aggregated by subject category (**Table 4** and **Figure 5**; exam-level distributions in **Supplementary Table 2**). Student performance is summarized as lowest and average normalized student grades per category, whereas model performance is computed on the SteuerEx benchmark, which aggregates questions from multiple examinations with heterogeneous grading schemes and point distributions. Because categories differ strongly in structure and total achievable points (**Table 4**), and exam difficulty varies across semesters (**Supplementary Table 2**), this comparison is descriptive and intended as contextual reference rather than direct human-model equivalence.

Across all six domains, average normalized student grades exceed model performance, with student category means ranging from 54% to 63% (**Table 4**). SteuerLLM performance ranges from 13% to 49%, and Small-SteuerLLM remains consistently lower. The largest gaps occur in categories that combine extensive statutory interpretation with procedural detail and computations. In corporate tax, students average 57%, while SteuerLLM achieves 36% (Small-SteuerLLM: 16%). In income tax, students average 63%, compared to 22% for SteuerLLM, reflecting the persistent difficulty of high-stakes income tax cases that demand both numerical accuracy and tightly structured legal justification.

At the same time, SteuerLLM does not uniformly fall below the full student performance range. It exceeds the lowest normalized student grade in corporate tax (36% vs. 1%), fiscal code (13% vs. 10%), fundamentals of tax law (49% vs. 10%), and VAT (32.5% vs. 17.6%) (**Table 4**). This indicates that, in several domains, the model can reach or surpass the lower tail of observed student outcomes and produces partially correct, exam-relevant reasoning rather than consistently failing responses. In contrast, taxation of partnerships remains difficult even relative to the weakest student results (SteuerLLM: 24% vs. lowest student grade: 46%), consistent with its narrow specialization and high per-question complexity. The closest alignment between model and student performance is observed in foundational material. In fundamentals of tax law, SteuerLLM reaches 49%, substantially improving over Small-SteuerLLM (32%) and reducing the gap to the student average (57%) relative to other categories. This pattern is consistent with SteuerLLM's training emphasis on broadly applicable statutory interpretation and reusable legal reasoning templates, whereas domains dominated by specialized edge cases and dense procedural constraints show larger remaining deficits.

**Table 4:** Composition of the SteuerEx benchmark and comparison of student and model performance across tax law categories. The table consolidates benchmark composition and performance metrics by tax law category. For each category, it reports the number of included examinations and participating students, the total number of examiner-defined statements, questions, and maximum achievable points in SteuerEx, as well as the semesters and exam names covered. Student performance is summarized by the lowest and average normalized exam grades, aggregated across the underlying examinations at Friedrich-Alexander-Universität Erlangen-Nürnberg. Corresponding normalized grades for Small-SteuerLLM and SteuerLLM are reported for the same categories. Categories are mutually exclusive, and each examination question contributes to exactly one category. Results are intended for contextual, category-level comparison rather than direct equivalence between individual student and model performance.

| Category | Corporate tax | Fiscal code | Fundamentals of tax law | Income tax | Taxation of partnerships | Value-added tax (VAT) |
|---|---|---|---|---|---|---|
| Total exams [n] | 5 | 2 | 5 | 3 | 1 | 2 |
| Total students [n] | 419 | 45 | 431 | 59 | 16 | 53 |
| Total statements [n] | 241 | 76 | 268 | 55 | 26 | 86 |
| Total maximum points | 261.5 | 129.0 | 269.0 | 189.0 | 66.0 | 121.0 |
| Total questions [n] | 44 | 3 | 56 | 4 | 4 | 4 |
| Included exam names | UnternehmenSt | AO | GrldStR | EStR | PersG | USt |
| Semesters covered | SS18, SS19, SS20, SS22, SS23 | SS20, WS16/17 | SS21, WS19/20, WS21/22, WS22/23, WS23/24 | WS19/20, WS20/21, WS21/22 | SS19 | SS21, SS22 |
| Lowest student grade [%] | 0.8 | 10.4 | 9.6 | 25.8 | 46.2 | 17.6 |
| Average student grade [%] | 56.9 | 54.2 | 56.9 | 63.3 | 60.2 | 55.9 |
| Small-SteuerLLM grade [%] | 15.7 | 6.7 | 32.1 | 13.4 | 15.2 | 27.8 |
| SteuerLLM grade [%] | 36.2 | 13.4 | 49.2 | 22.4 | 23.5 | 32.5 |

The student comparison suggests that SteuerLLM does not reach average student performance across any category, but demonstrates non-trivial competence relative to the lowest observed student outcomes in multiple domains. Given the aggregated nature of the student data and the heterogeneous exam composition underlying SteuerEx, these results should be interpreted as contextual benchmarks that situate model performance within a realistic academic assessment setting rather than as a direct ranking against individual students.

## Human expert evaluation and validation of automated grading

To assess the validity and robustness of the automated statement-level evaluation used throughout this study, we conducted an additional human evaluation on a stratified subset of model outputs (**Supplementary Table 3**). This analysis focuses on two complementary aspects: the consistency of human grading at the statement level and the agreement between human judgments and the automated LLM-based evaluator. Human grading of tax law examination answers at the statement level exhibits substantial variability. Across 20 statements independently graded by three evaluators with tax law background, inter-rater reliability was low (ICC(2,1) = 0.367). Perfect agreement among all three raters occurred in only 3 out of 20 cases. Pairwise correlations further indicate uneven agreement patterns, with moderate correlation between two evaluators and weak, non-significant correlations involving the third. These results highlight that fine-grained partial-credit grading in tax law is inherently subjective, particularly for statements that are only partially satisfied. This variability mirrors real examination settings, where grading discretion plays a substantial role, and underscores the difficulty of defining a single, authoritative ground truth at this level of granularity.

Despite this variability among human graders, the automated evaluator shows strong alignment with aggregated human judgment. Across 59 statement-level evaluations with valid human and automated scores, the rank correlation between averaged human scores and LLM-assigned scores was high (Kendall's τ = 0.718, 95% CI [0.599, 0.820]). This strong correlation indicates that, although individual human assessments differ, the automated evaluation captures the central tendency of expert judgment reliably. Importantly, this agreement is observed in precisely those cases that are most diagnostically informative, namely statements receiving partial credit rather than clear failures or fully correct answers. Model-specific analyses reveal consistently strong correlations between human and automated scores across all evaluated systems. Kendall's τ ranges from 0.730 to 0.756 for DeepSeek-R1-671B, Llama-3.2-3B-it, SteuerLLM, and GPT-4o-mini, with all correlations statistically significant despite modest sample sizes per model. This consistency suggests that the automated evaluator does not favor a particular architecture or training paradigm, but applies a stable grading standard across heterogeneous models.

Consequently, these findings support the use of automated statement-level evaluation for comparative benchmarking in SteuerEx. While human graders disagree substantially at the individual statement level, the automated evaluator aligns closely with the aggregated human signal and provides a reproducible, scalable alternative to manual grading. The goal of SteuerEx is not to replicate individual examiner decisions, but to enable consistent relative comparison of model performance across a large and diverse set of authentic tax law questions. In this context, the observed human-LLM agreement provides empirical justification for the evaluation methodology employed throughout this study. A representative example of the human grading interface and statement-level assessment process is shown in **Supplementary Figure 1**.

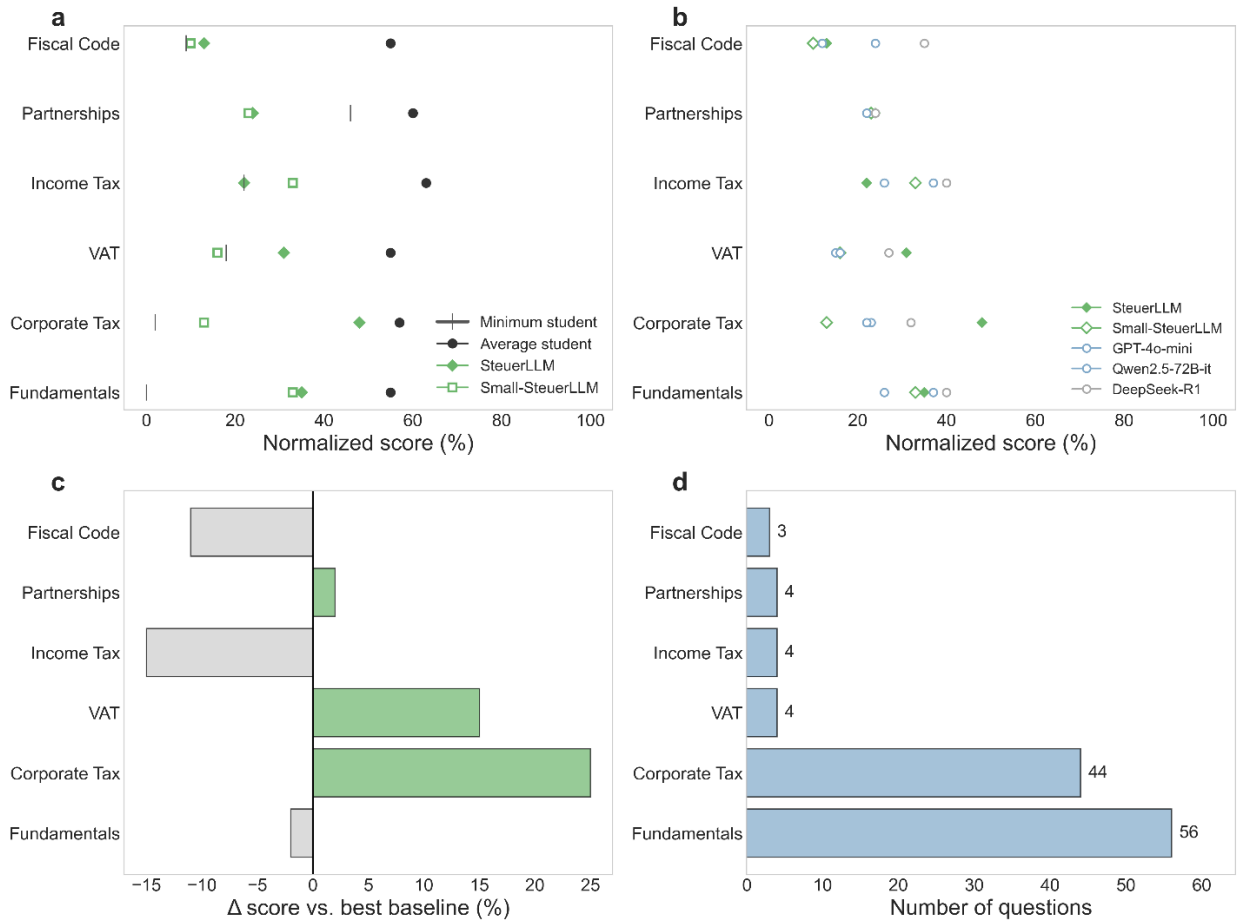

**Figure 5**: Category-level performance across tax law domains on SteuerEx. **a** Average normalized scores for students and domain-adapted models by tax law category, showing the lowest observed student score, average student score, Small-SteuerLLM, and SteuerLLM. Student scores are aggregated across all available examinations per category. **b** Category-level normalized scores for selected language models, including general-purpose, reasoning-oriented, and domain-adapted systems, evaluated on the corresponding subsets of SteuerEx. **c** Difference in normalized score between SteuerLLM and the strongest non-tax-specialized baseline per category (Δ score), with positive values indicating higher performance of SteuerLLM. **d** Number of SteuerEx questions per tax law category. Model scores are computed on the benchmark questions only, and categories differ in size and point weighting, precluding direct aggregation across domains.

# Open-SteuerLLM: impact of removing private training data

To enable public release of model weights and training details, we trained Open-SteuerLLM, a variant of SteuerLLM that uses the same architecture, optimization procedure, and synthetic data generation pipeline, but excludes a private subset of the training data that cannot be redistributed. This comparison isolates the effect of reduced training data while holding all other factors constant.

On the full SteuerEx benchmark (**Table 2**), Open-SteuerLLM achieves 23% ± 3 (95% CI: 18–29; 241/1,035.5 points), compared to 28% ± 2 for SteuerLLM (294/1,035.5 points). The mean performance of the open model is therefore lower in absolute terms, corresponding to a reduction of approximately 53 points (≈5 percentage points). However, this difference is not statistically significant under paired bootstrap testing (P = 0.19), indicating that the observed gap cannot be distinguished from sampling variability at the benchmark level.

Category-level results (**Table 3**) show that the effect of removing private training data is uneven across domains. In corporate tax, the largest and most heavily weighted category (44 questions), Open-SteuerLLM matches SteuerLLM exactly at 38% ± 4, indicating no measurable loss in this core domain. In fundamentals of tax law (56 questions), Open-SteuerLLM slightly exceeds SteuerLLM (42% ± 3 vs. 41% ± 3), well within confidence intervals. Income tax performance is likewise identical (24% ± 1 for both models). In contrast, Open-SteuerLLM performs worse in smaller categories. In fiscal code (3 questions), its score drops from 14% ± 0 to 6% ± 0, and in value-added tax from 35% ± 14 to 22% ± 5. A similar reduction is observed in taxation of partnerships (18% ± 8 vs. 11% ± 8). These domains have very few questions, and category scores are therefore highly sensitive to individual items; nonetheless, the direction of the difference is consistent and suggests that the removed private data provided additional coverage for specialized or edge-case material.

Overall, Open-SteuerLLM preserves the qualitative performance profile of SteuerLLM and remains competitive with strong general-purpose baselines, but exhibits a moderate absolute performance reduction relative to the full model. The largest domains, which dominate the overall benchmark score, are largely unaffected, while losses are concentrated in narrowly sampled categories with high per-question complexity. These results indicate that the majority of SteuerLLM's gains arise from the publicly reproducible training pipeline and synthetic data generation strategy, while the private data contribute incremental improvements, particularly in specialized subdomains. By releasing Open-SteuerLLM, we provide the research community with a fully open-weight tax-law-specialized model whose performance remains close to the closed variant, while clearly documenting the performance trade-offs introduced by strict data openness.

# Discussion

In this study, we introduce two tightly coupled contributions: SteuerEx, the first open benchmark derived from authentic German university tax law examinations, and SteuerLLM, a domain-adapted large language model specifically trained for German tax law reasoning. SteuerLLM is trained on a large-scale synthetic dataset generated from authentic examination material using a controlled retrieval-augmented pipeline and a block expansion strategy that adds domain-specific capacity while largely preserving pretrained representations. Using SteuerEx, we evaluate a broad range of instruction-tuned and reasoning-oriented LLMs spanning 3B to 671B parameters under a statement-level, partial-credit grading scheme that mirrors real academic examinations. Across this diverse comparison, we find that authentic tax law exams remain highly challenging for current models, and that targeted domain adaptation, as implemented in SteuerLLM, yields substantial and statistically robust gains over general-purpose instruction tuning. At the same time, performance improvements are uneven across tax law domains, and several categories remain difficult even for specialized models, underscoring that highly structured legal reasoning continues to pose fundamental challenges for modern LLMs. Beyond introducing a new benchmark, this work demonstrates that realistic academic assessment formats can expose capability differences that are largely obscured by conventional legal NLP benchmarks[11,16–18]. By operationalizing tax law evaluation through structured reasoning, statutory precision, and incremental partial credit, SteuerEx provides a more faithful measure of model generalization to real-world legal reasoning and clarifies why models that perform well on generic reasoning tasks can still fail under examination-style constraints.

A central finding of this study is that performance on SteuerEx does not scale monotonically with parameter count across model families. Although the strongest overall system in our comparison is DeepSeek-R1-671B several substantially larger instruction-tuned models perform well below SteuerLLM, with 28B parameters, despite having up to an order of magnitude more parameters. Conversely, Small-SteuerLLM, with only 10B parameters, performs competitively with mid-sized general-purpose models in the 14B–32B range. These results indicate that, for German tax law analysis, training data composition and alignment with domain-specific answer structure are more decisive than raw model scale[12]. In this setting, general capacity increases alone appear insufficient to induce reliable gains. Instead, models benefit from exposure to tax-specific reasoning patterns, precise statutory citation practices, and structured legal argumentation that directly correspond to examination grading rubrics. This finding is consistent with prior observations that scaling laws derived from general benchmarks do not necessarily transfer to domains with strict formal constraints and specialized evaluation criteria[27,28].

The category-level analysis further clarifies where domain adaptation is most effective and where its limits become apparent. SteuerLLM shows its largest and most stable advantages in high-coverage domains such as corporate tax and fundamentals of tax law, matching or closely approaching the strongest non-specialized baselines. These categories comprise the majority of questions in SteuerEx and emphasize multi-step statutory subsumption, hierarchical norm interpretation, and structured written reasoning, all of which align closely with SteuerLLM's training objectives. In contrast, performance is lower and more volatile in narrowly sampled domains, including fiscal code, income tax, taxation of partnerships, and VAT. In these categories, a small number of high-weight questions can dominate outcomes, and rare procedural edge cases may remain underrepresented even in a large synthetic training corpus. This pattern highlights an important limitation of domain adaptation: while targeted training can substantially improve performance in broad doctrinal areas, it does not automatically resolve sparsely sampled subdomains where exceptional rules, procedural nuance, or high point density amplify the cost of individual reasoning errors.

Comparison with anonymized student examination outcomes provides an external reference point for interpreting model performance under realistic academic conditions. Across all six tax law categories, average student scores exceed all LLMs, confirming that current LLMs do not meet typical university examination standards in German tax law. At the same time, SteuerLLM exceeds the lowest observed student outcomes in several domains, including corporate tax, fiscal code, fundamentals of tax law, and VAT. This overlap indicates that the model can produce partially correct, exam-relevant legal reasoning that attains non-trivial credit under authentic grading schemes, rather than merely generating superficially plausible answers. Nonetheless, the persistent gap to average student performance, and the particularly pronounced deficit in taxation of partnerships, underscore that exam-level tax law reasoning remains unsolved for current models. These shortcomings are most evident in domains that require dense integration of statutory doctrine, procedural constraints, and numerical computation, where partial errors can invalidate large portions of an answer.

A central methodological challenge in legal benchmarking is whether automated evaluation remains reliable when grading depends on partial credit and nuanced assessments of legal equivalence. Our human evaluation confirms that statement-level grading in tax law is itself subjective: inter-rater reliability among three expert evaluators was low, with perfect agreement occurring in only a small subset of overlap items. Despite this variability, the automated LLM-based evaluator exhibits strong alignment with aggregated human judgment, consistently across multiple evaluated models. This result supports the use of automated statement-level grading for comparative benchmarking at scale, particularly when the goal is to capture relative performance differences rather than replicate individual examiner decisions[29]. At the same time, the observed human disagreement highlights an important property of realistic legal assessment: at the level of granularity required to mirror examination grading, no single authoritative ground truth exists. SteuerEx therefore prioritizes reproducibility, transparency, and consistency over emulating individual grading discretion, providing a stable basis for systematic model comparison under authentic assessment conditions.

The comparison between SteuerLLM and Open-SteuerLLM illustrates the trade-off between data openness and performance in domain-specialized language models. Open-SteuerLLM uses the same architecture, optimization strategy, and synthetic data generation pipeline as SteuerLLM, differing only by the exclusion of a private subset of training data that cannot be redistributed. On the full SteuerEx benchmark, Open-SteuerLLM attains a lower mean score than SteuerLLM, but this difference is not statistically significant. At the category level, performance is largely preserved in the largest and most heavily weighted domains: Open-SteuerLLM matches SteuerLLM exactly in corporate tax and shows no meaningful difference in fundamentals of tax law or income tax. Performance reductions are concentrated in smaller categories, including fiscal code, VAT, and taxation of partnerships, where individual questions carry high point weight and category-level estimates are therefore sensitive to limited sample size. The consistent direction of these differences suggests that the removed private data contributed incremental coverage for specialized or edge-case material rather than driving overall benchmark performance[30]. These results indicate that the majority of SteuerLLM's gains arise from the publicly reproducible training pipeline and large-scale synthetic data generation strategy, enabling an open-weight release that remains close in performance to the closed model while supporting transparency and reproducibility[30,31].

This study has several limitations. First, although SteuerEx is derived from authentic German university tax law examinations spanning multiple years and both bachelor and master levels, all source material originates from a single academic institution. While German tax law is nationally standardized at the statutory level, examination design, grading emphasis, and stylistic conventions vary across universities and instructors. As a result, absolute performance levels and some domain-specific patterns observed on SteuerEx may not fully generalize to other institutional settings. This institutional concentration also propagates into the synthetic training data used for SteuerLLM, which is ultimately anchored in the distribution of the original seed examinations and the legal sources retrieved during generation. Consequently, rare doctrinal edge cases, unusual procedural constellations, or highly specialized statutory exceptions may remain underrepresented despite large-scale synthesis[32]. Expanding the benchmark with examinations from additional institutions and augmenting the seed material for sparsely covered subdomains will be necessary to further improve robustness and generalization in complex areas of tax law[12,33]. Second, uncertainty estimates depend on the resampling design. Because questions differ substantially in maximum point value, we used a points-constrained bootstrap that samples questions with replacement until a fixed target total of maximum points is reached. This exact-sum constraint may change the effective inclusion probabilities of questions, as low-point questions are more likely to be selected in the final steps needed to match the target. As a result, bootstrap variance estimates and confidence intervals may differ from those obtained under standard question-level resampling and may not always align with permutation-based p values (for more details, see **Supplementary Note 1)**. Third, the distribution of questions across tax law domains is highly uneven. Corporate tax and fundamentals of tax law account for the majority of benchmark questions and points, whereas other categories, including fiscal code, income tax, VAT, and taxation of partnerships, contain only three to four questions each. As a result, category-level estimates in these domains are statistically unstable and sensitive to individual high-weight questions, limiting the reliability of fine-grained domain comparisons and error attribution. Future work could expand SteuerEx with additional authentic examinations, particularly in sparsely represented domains. Fourth, SteuerEx is intentionally restricted to text-only examination content. Examinations that rely predominantly on tables, graphical elements, structured forms, or interactive calculations were excluded to ensure compatibility with text-based language models and controlled evaluation. While this design choice improves reproducibility, it omits substantial components of real tax law practice and assessment, where structured artifacts and formal calculation templates play a central role. Benchmark performance should therefore not be interpreted as a comprehensive measure of real-world tax advisory competence[33,34]. Fifth, models were evaluated without retrieval augmentation[10,35], external legal databases[36], or access to current statutory texts at inference time. This isolates internalized tax law knowledge and exam-style reasoning, but does not reflect realistic professional workflows in which retrieval and citation verification are essential, especially in a legal domain characterized by frequent statutory amendments. The remaining performance gap may therefore overstate limitations of deployed systems that integrate retrieval-based grounding[3]. Sixth, the evaluation relies on automated statement-level grading using an external LLM evaluator. Although human validation shows strong alignment with aggregated expert judgment, statement-level grading in tax law is inherently subjective, as reflected by low inter-rater reliability among human evaluators. Automated scores should therefore be interpreted as comparative and relative rather than definitive measures of legal correctness, particularly for partially correct reasoning.

Overall, this work establishes a rigorous and realistic framework for evaluating tax law reasoning in language models. By combining an authentic exam-based benchmark with partial-credit scoring, a broad multi-model evaluation, and an openly released domain-specialized model, we provide both a diagnostic tool and a concrete baseline for future progress. The results suggest that domain-specific data and targeted architectural adaptation can materially improve performance in structured legal reasoning tasks, but also that substantial gaps remain relative to human examinees, especially in narrowly sampled and highly specialized domains. More broadly, SteuerEx demonstrates how real academic assessments can serve as high-fidelity benchmarks for studying specialization, generalization, and scaling behavior in deep learning systems, and it offers an open foundation for building more reliable and transparent legal AI.

# Methods

## Ethics statement

This study was conducted in accordance with applicable ethical standards and institutional regulations. The SteuerEx benchmark is derived from past university tax law examinations for which the authors had legitimate access and permission for research use. All student examination results were fully anonymized prior to analysis, aggregated at the category level, and contained no personally identifiable information. No interaction with students occurred, and no individual-level data were analyzed. As the study does not involve human subjects research within the meaning of applicable regulations, institutional review board approval and informed consent were not required.

## SteuerEx benchmark

To enable a rigorous and realistic evaluation of LLMs in the domain of German tax law, we introduce SteuerEx, a benchmark derived from authentic university tax law examinations administered at German academic institutions. SteuerEx consists of 115 examination questions with a total achievable score of 1,035.5 points, reflecting the weighted grading schemes used in real academic assessments. Each question is paired with an expert-validated reference solution and a detailed scoring rubric that decomposes the solution into graded legal statements, enabling fine-grained and reproducible evaluation. To the best of our knowledge, SteuerEx is the first openly available benchmark specifically designed to assess large language models on German tax law reasoning. The dataset originates from original tax law examinations conducted by the Chair of Tax Law and Public Law at Friedrich-Alexander-Universität Erlangen-Nürnberg (FAU). The source material includes both bachelor-level examinations, which emphasize foundational doctrinal knowledge, and master-level examinations, which focus on complex case-based reasoning requiring advanced statutory interpretation and structured legal argumentation. Only examinations from 2016 onward were considered, with a deliberate emphasis on examinations from 2020 and later, ensuring alignment with the current German tax law framework.

To ensure compatibility with text-based model evaluation, examinations relying predominantly on non-textual materials such as extensive tables, calendars, or graphical elements were excluded. Where limited structured elements were present, they were standardized into Markdown format to preserve informational content. Questions containing multiple subparts were decomposed into individual items while preserving the full factual and legal context. After preprocessing, the final benchmark comprises 115 questions drawn from 18 distinct examinations and is provided in a structured JSON format to support systematic evaluation and reproducibility.

SteuerEx covers a broad and representative spectrum of German tax law and comprises six disjoint tax law categories derived from authentic university examinations: Unternehmensbesteuerung (*UnternehmenSt*; corporate tax), Abgabenordnung (*AO*; fiscal code and tax procedure), Grundlagen des Steuerrechts (*GrldStR*; fundamentals of tax law), Einkommensteuerrecht (*EStR*; income tax law), Besteuerung von Personengesellschaften (*PersG*; taxation of partnerships), and Umsatzsteuerrecht (*USt*; value-added tax). Each category corresponds to one or more complete exam-semester pairs and reflects established curricular divisions in German tax law education. All benchmark questions are assigned to exactly one category, ensuring mutually exclusive coverage across domains. The detailed composition of the benchmark by category is reported in **Table 4**.

Each reference solution in SteuerEx is decomposed into discrete legal statements, each assigned a point value reflecting its relevance to correct legal classification, statutory citation, procedural reasoning, or numerical accuracy. Statement weights range from 0.5 to 13 points, consistent with real examination grading practices. Model-generated answers are evaluated against these statements based on conceptual correctness, factual accuracy, and precision of statutory references, with partial credit awarded where legally appropriate. This design mirrors the structure of German tax law examinations, which require structured legal writing, precise statutory grounding, and the integration of multiple interdependent legal norms rather than factual recall alone. All benchmark questions, reference solutions, and scoring schemes were reviewed and validated by tax law experts affiliated with FAU, with expertise spanning income taxation, corporate taxation, and fiscal procedure. Benchmark construction and supervision were conducted by Q.J. at the Chair of Tax Law and Public Law at the School of Business, Economics and Society of FAU, who has been affiliated with the chair since January 2020. This expert validation ensures that SteuerEx accurately reflects real examination standards and professional expectations and provides a realistic evaluation setting for assessing whether models trained on curated or synthetic legal data generalize to authentic academic assessment tasks.

## Algorithmic generation of the training dataset

The limited availability of large-scale, structured, and publicly accessible datasets in German tax law necessitated the development of a dedicated synthetic data generation pipeline for training SteuerLLM. The pipeline is grounded in authoritative legal sources, including statutory texts such as the Einkommensteuergesetz and the Abgabenordnung, supplemented by administrative guidelines and legal commentaries. These sources form the legal foundation for generating domain-consistent training data.

In collaboration with professional tax advisors, we defined an extensive taxonomy of tax-relevant question types designed to reflect real-world advisory and examination scenarios. The framework includes classification tasks, procedural ordering problems, complex scenario-based legal reasoning, and detailed numerical computations required for tax declarations and deductions. A total of 18 distinct question types were specified, each accompanied by formal generation instructions and illustrative examples (see **Supplementary Note 2**). These specifications guided the automated generation process and ensured consistency in structure, content, and complexity (see **Supplementary Note 3**).

The core of the synthetic data generation process is the Water Fountain Algorithm, a custom-designed iterative pipeline that automatically generates realistic and legally precise question-answer pairs without manual intervention. The full algorithm is provided in **Supplementary Note 4**, with a schematic overview of the processing steps shown in **Supplementary Figure 1**. The algorithm is initialized with a curated seed dataset consisting of approximately 1,800 authentic tax law examination questions from FAU. These seed questions were selected to ensure broad topical coverage, as the diversity of the initial set directly influences the variability and comprehensiveness of the generated dataset.

Each seed question is first transformed into a search query and submitted to a locally hosted instance of the SearXNG meta-search engine. The search retrieves relevant documents from online German tax law resources. Retrieved documents are segmented into textual chunks, and each chunk is evaluated for semantic relevance to the original question using cosine similarity based on vector[37]:

$$\text{sim}(c, q) = \frac{c \cdot q}{|c||q|} \quad (1)$$

Here, c and q denote the vector embeddings of the chunk and the question, respectively. Embeddings are generated using the multilingual model intfloat/multilingual-e5-large[5]. Chunks are sorted in descending order of semantic similarity[38]. To construct the model prompt, chunks are sequentially aggregated until the model's context window limit N is reached, such that

$$\sum_{i=1}^{k} tokens(c_i) \leq N \quad \text{and} \quad \sum_{i=1}^{k+1} tokens(c_i) > N. \tag{2}$$

If adding a chunk exceeds the token limit, the chunk is truncated accordingly. The resulting context $C_q^*$ and the corresponding question $q$ are passed to the language model to generate an answer $A_q$:

$$G(C_q^*, q) \rightarrow A_q. \tag{3}$$

The model is explicitly instructed to output a predefined flag if the provided context is insufficient. This behavior is formalized by the binary indicator function:

$$f(A_q) = \begin{cases} 1, & \text{if context is insufficient} \\ 0, & \text{otherwise} \end{cases} \tag{4}$$

All instances for which $f(A_q) = 1$ are discarded to maintain dataset quality. For each validated question-answer pair, the algorithm generates multiple new, thematically diversified questions based on the existing context:

$$G: Q_n \times C_q^* \rightarrow Q_{n+1}. \tag{5}$$

Each answered question generates three additional questions, which are fed back into the next iteration. The dataset size at iteration n is therefore given by:

$$|Q_n| = k \cdot |Q_{n-1}| \tag{6}$$

where $Q_0$ denotes the initial seed set and k is the growth factor, set to k = 3 in our implementation. This process ensures controlled exponential growth while continuously expanding topical coverage beyond the original seed questions.

## Data cleansing and exclusion criteria

Following generation, a multi-stage cleansing process was applied to ensure dataset integrity and quality. In the first stage, all generated question-answer pairs were checked for exact duplicates and overlaps with the initial seed dataset. A total of 47,555 duplicated or overlapping tuples were removed to maintain strict separation between authentic examination material and synthetic data. In the second stage, all tuples associated with explicitly flagged generation errors were excluded. During generation, the model was instructed to return a predefined error string whenever it encountered insufficient context or failed generation. This step resulted in the removal of an additional 11,483 tuples. In a third stage, all question-answer pairs containing partial occurrences of the error flag were eliminated, as these were interpreted as indicators of incomplete or unreliable generation.

Additionally, the sufficiency of external sources retrieved during the retrieval-augmented generation process was evaluated. A minimum of three independent retrieved documents was required for an instance to be retained. Tuples based on fewer sources were excluded due to the increased risk of contextually inaccurate answers. Across all stages, 120,625 tuples were removed (see **Supplementary Table 4**). After cleansing, the final dataset consisted of 485,092 validated question-answer pairs derived from an initial set of 605,717 generated instances.

After quality control, the dataset was structured and categorized to support targeted model training. The final dataset comprises two principal components. The complete composition and categorization of the final training dataset are summarized in **Supplementary Table 5**. The primary synthetic generation component includes the majority of examples produced directly through the Water Fountain Algorithm and spans a wide range of advisory and examination-style tasks. Within this component, several subgroups were defined to increase topical diversity, emphasize statutory interpretation, and improve numerical reasoning. In addition, a context-supported generation component was constructed by reusing validated legal contexts from the primary dataset to generate additional question-answer pairs closely aligned with previously verified material. This approach improves contextual consistency while expanding dataset size.

In total, the final dataset comprises 485,092 validated question-answer pairs (see **Supplementary Table 5**). Together with the SteuerEx benchmark, this dataset forms one of the largest and most structured resources for German tax law reasoning and is publicly released to support reproducibility and further research in domain-specific legal artificial intelligence.

## Model extension and training

SteuerLLM is derived from pretrained decoder-only Transformer[2] models, using Mistral Small 2409 (24B parameters) and Ministral (8B parameters) as base architectures. The design objective is to incorporate domain-specific knowledge for German tax law while preserving the general-purpose reasoning and language capabilities of the original models. To achieve this, we adopt a block expansion strategy in which additional Transformer layers are introduced and trained while the original pretrained layers remain frozen.

Starting from the 24B Mistral Small model, we expand the network to 28B parameters by inserting eight additional Transformer blocks into the depth of the model. An analogous expansion is applied to Ministral, increasing it from 8B to 10B parameters. The new blocks are interleaved between existing layers rather than appended at the output, allowing domain-specific representations to emerge at multiple levels of abstraction while maintaining the internal feature organization of the base model. Each inserted block is architecturally identical to the original layers, including multi-head self-attention, gated feed-forward networks, and layer normalization, ensuring full compatibility with the pretrained inference stack.

To stabilize training, newly inserted blocks are initialized from adjacent pretrained layers[39]. Selected projections within these blocks are initialized to zero, such that the residual pathways initially approximate identity mappings. This initialization ensures that, prior to training, the expanded model behaves similarly to the base model, and domain-specific capacity is gradually activated during optimization rather than disrupting pretrained representations[40].

Training is performed using selective unfreezing[41]. All original Transformer blocks remain frozen throughout training, while only the newly inserted blocks are updated. In addition, the token embedding matrix and the language modeling head are trained to accommodate new domain-specific terminology and output distributions. This approach concentrates learning in the expanded capacity and minimizes catastrophic forgetting of general language and reasoning skills[42]. The model is trained in two stages. First, continued pretraining is performed on 6.3 billion tokens of domain-relevant German text, including filtered German web sources. Second, instruction tuning is applied using approximately 1.5 million tax-domain instruction-response pairs, aligning the model with structured reasoning and assistant-style outputs typical of tax advisory and examination settings[19,20].

Training optimizes the standard causal language modeling objective[43]. Given a tokenized sequence $\{x_t\}_{t=1}^{T}$, the loss is:

$$\mathcal{L} = -\sum_{t=1}^{T} \log p_\theta(x_t \mid x_{<t}) \tag{7}$$

with teacher forcing. For instruction data, we apply response-only training[8] (i.e., do not backpropagate through the prompt tokens), ensuring gradients primarily shape the assistant's completions while maintaining prompt semantics.

Optimization is performed using AdamW[44] with cosine learning-rate decay and warmup. Training uses sequence packing and mixed-precision arithmetic to maximize throughput[45]. Distributed training is carried out across multiple nodes equipped with NVIDIA H100 GPUs using parameter sharding, enabling efficient scaling while keeping the trainable parameter subset memory-resident[46]. By allocating new depth-wise capacity while freezing pretrained parameters, this block expansion strategy enables SteuerLLM to acquire high-precision tax-domain knowledge without degrading its general-purpose capabilities. This design allows the model to specialize in statutory interpretation, legal phrasing, and numerical tax reasoning while remaining efficient enough to run on modest hardware.

# Experimental design

All evaluation questions in this study are drawn from authentic German university tax law examinations administered between the winter semester 2016/17 (WS16/17) and the winter semester 2023/24 (WS23/24) at FAU. The benchmark covers six core subject areas that recur across the underlying examinations: fundamentals of tax law (Grundlagen des Steuerrechts)[41], corporate tax (Unternehmenssteuerrecht)[47], VAT (Mehrwertsteuer), income tax (Einkommensteuer), taxation of partnerships (Personengesellschaften), and the fiscal code (Abgabenordnung). In total, the benchmark comprises 115 examination questions with examiner-provided reference solutions and a maximum achievable score of 1,035.5 points. The distribution of questions, statements, and maximum points across subject areas, as well as the covered examinations and semesters, is reported in **Table 4**.

The central design objective of the benchmark evaluation is to approximate real academic grading while enabling scalable and reproducible model comparisons. To this end, each reference solution is decomposed into atomic, independently scorable legal statements ("pointable statements")[48]. Across all 115 questions, this results in 752 statements. Each statement is assigned a maximum point value $m_{q,i}$ that reflects its relevance under the original grading scheme. Statement weights vary across questions and domains, preserving examiner emphasis on key legal qualifications, required statutory citations, procedural steps, and numerical sub-results. This statement-level representation allows partial credit and avoids forcing binary correctness on multi-part legal reasoning tasks.

The evaluation pipeline consists of two stages: (i) model answering and (ii) statement-wise grading. In the answering stage, each evaluated model receives only the original exam question text, exactly as presented to students, including the full factual narrative and sub-questions where applicable[49]. Models do not receive the reference solution, statement annotations, examples, chain-of-thought scaffolding[4], retrieval augmentation[10], or external legal context. This isolates a model's ability to produce an exam-style response from the question alone. All model runs were executed with deterministic decoding[50] (temperature = 0) to ensure reproducibility at the response level and to prevent variability due to sampling. The evaluation included 17 LLMs: DeepSeek-R1-Distill-Llama-70B, Llama-3.2-3B-it[51,52], Llama-3-8B-it[51,52], Ministral-8B-it-2410, Mistral-Small-it-2409, Qwen2.5-14B-it[53], Qwen2.5-32B-it[53], Qwen2.5-3B-it[53], Qwen2.5-72B-it[53], Qwen2.5-7B-it[53], Small-SteuerLLM, SteuerLLM, Open-SteuerLLM, DeepSeek-R1-671B[54], Gemma-3-27B-it[55,56], Gemma-3-4B-it[55,56], and GPT-4o-mini[54]. Because evaluated models differ substantially in native context length and generation defaults (**Table 1**), we enforced a uniform output budget to maintain comparability. For all models, we imposed a global cap of max_tokens = 4096 for answer generation. This cap prevents uncontrolled expansion toward a model's maximum context window and ensures that differences in performance are not driven by differences in answer length allowances. Reasoning-oriented models can emit long intermediate traces that contain exploratory or contradictory hypotheses that are not intended as part of a submitted exam answer and could artificially increase apparent statement coverage during grading. For models that expose such traces, we removed the reasoning trace prior to grading and retained only the final answer section[4,57]. In rare cases, the global max_tokens = 4096 cap caused a reasoning-oriented model to exhaust the token budget within the trace and produce no final answer[58]. To avoid systematic under-scoring due to truncation artifacts, we increased the output cap specifically for DeepSeek-R1-671B to max_tokens = 32768, ensuring that a final answer was produced after internal reasoning. As in all other cases, only the final answer text (with traces removed) was passed to the grading stage.

In the grading stage, each model answer is scored against the examiner solution at the statement level using an external LLM evaluator (GPT-4o, deployed via Azure OpenAI). Using a fixed external evaluator ensures that no evaluated model grades its own outputs and that all candidate models are judged under a single, consistent standard. For a given question $q$, let $\mathcal{S}_q = \{1, \ldots, n_q\}$ denote the set of statements in its reference solution. Statement $i \in \mathcal{S}_q$ has a maximum point value $m_{q,i} > 0$. The evaluator assigns awarded points $a_{q,i}$ constrained to the closed interval:

$$0 \leq a_{q,i} \leq m_{q,i} \tag{8}$$

The evaluator is instructed to score based on semantic and legal equivalence rather than lexical overlap[59]. Full credit is awarded if the substantive legal content of the statement is correctly represented in the model answer, partial credit if the statement is only partially correct or incomplete (for example correct qualification but missing a required statutory citation, or correct citation but incomplete reasoning), and zero if it is missing or incorrect. The evaluator is explicitly provided the statement's maximum point value $m_{q,i}$ to calibrate partial-credit decisions. This is necessary because statements differ substantially in weight and complexity and partial credit should scale appropriately[49].

The evaluation prompt is constructed per $(q, i)$ pair and includes: (i) the original question text, (ii) the full gold reference solution for the question, (iii) the model-generated answer (with any reasoning traces removed), (iv) the statement identifier and statement text to be graded, and (v) the maximum points $m_{q,i}$. The evaluator is constrained to output a single valid JSON object containing the awarded points $a_{q,i}$, the maximum points $m_{q,i}$, the statement identifier, and a one-sentence justification. Enforcing a strict JSON-only output format enables robust automated parsing, aggregation, and auditing at scale. Scores are aggregated to mirror examination grading while enabling comparisons across models and domains. The raw score for question $q$ is:

$$A_q = \sum_{i=1}^{n_q} a_{q,i}, \tag{9}$$

and the maximum achievable score for that question is:

$$M_q = \sum_{i=1}^{n_q} m_{q,i}. \tag{10}$$

Across the full benchmark with $Q = 115$ questions, the total raw score is:

$$A_{\text{total}} = \sum_{q=1}^{Q} A_q, \tag{11}$$

and the benchmark maximum is:

$$M_{\text{total}} = \sum_{q=1}^{Q} M_q = 1035.5. \tag{12}$$

The primary evaluation metric reported throughout the paper is the normalized benchmark score expressed as a percentage of the benchmark maximum:

$$\text{Score}_\% = 100 \times \frac{A_{\text{total}}}{M_{\text{total}}}. \tag{13}$$

This normalized score is the basis for the headline results in **Table 2** and allows direct comparison across models despite heterogeneous distributions of question weights[23]. Category-level scores are computed analogously by restricting aggregation to the subset of questions belonging to a category $c$[11]. Let $Q_c$ denote the question indices in category $c$. Then:

$$A_c = \sum_{q \in Q_c} A_q, \tag{14}$$

$$M_c = \sum_{q \in Q_c} M_q, \tag{15}$$

$$\text{Score}_\%(c) = 100 \times \frac{A_c}{M_c}. \tag{16}$$

This preserves examiner weighting within each domain and supports category-level comparisons with aggregated student outcomes (**Table 4**). For comparisons with student examination outcomes, additional normalization was required to account for exam components that could not be evaluated by text-only language models, such as questions relying on figures, tables, or graphical completion. For these questions, the normalization reference was set to the higher of the model-derived maximum score and the highest observed student score, ensuring that student performance was not systematically penalized by modality-dependent components absent from the model evaluation. Model scores for such questions remained zero, reflecting the absence of evaluable output rather than incorrect responses. This procedure preserves the original grading structure while enabling fair aggregation of student and model scores[60,61].

All answering prompts were standardized to contain only the raw exam question text, and all grading prompts followed the same structured format across all statements and models. Deterministic decoding (temperature $= 0$) was used for answer generation, and grading outputs were constrained to machine-readable JSON. Together with the statement-wise decomposition and explicit maximum point values, this design ensures that scoring is reproducible, supports auditing at the level of individual legal requirements, and enables downstream analyses by category, examination year, and statement type. A schematic illustration of the full answer-and-grade pipeline and a worked example are provided in **Figure 2** for interpretability.

## Human evaluation and inter-rater reliability analysis

To validate the reliability of the automated LLM-based grading framework used throughout this study, we conducted an independent human evaluation and inter-rater reliability analysis on a carefully selected subset of model outputs[62]. The objectives were twofold: first, to assess agreement between human expert judgments and the automated evaluator, and second, to quantify the internal consistency of human grading at the statement level under realistic examination conditions.

From the full SteuerEx benchmark, we constructed a focused evaluation subset using deterministic stratified sampling[63]. Sampling was performed at the question level based on each model's normalized score relative to the maximum achievable points for that question. For each evaluated model, questions were ranked by percentage score and partitioned into three equally sized strata representing low, medium, and high model performance. From each stratum, questions were sampled uniformly at random using a fixed random seed to ensure reproducibility. This design ensured coverage of the full spectrum of model behavior rather than concentrating on trivial zero-score or near-perfect cases. Each sampled question was decomposed into its constituent statement-level grading units, matching the granularity of the automated evaluation pipeline. To maximize diagnostic value, statements receiving partial credit from the automated evaluator were preferentially selected. Specifically, statements with awarded scores between 5% and 95% of the maximum possible points were oversampled, as these cases are most informative for assessing ambiguity, borderline correctness, and grading subjectivity. Statements with near-zero or near-perfect automated scores were included only as needed to reach the target sample size. The final human evaluation set comprised 101 unique statement-level items drawn from 28 distinct examination questions. The evaluated models included DeepSeek-R1-671B, Llama-3.2-3B-Instruct, SteuerLLM, and GPT-4o-mini. Three independent human evaluators with tax law expertise (denoted HIWI_1, HIWI_2, and HIWI_3) participated in the study. HIWI_1 has seven years of professional experience at DATEV eG, a German service provider for tax advisors, where they worked on projects combining artificial intelligence with applications in German tax law, including automated tax analysis and compliance-related systems. HIWI_3 has several years of professional experience at DATEV eG in the company's AI-focused unit, contributing to the development of AI-based tools supporting tax advisory workflows and interpretation of tax regulations. HIWI_2 (L.S.) is affiliated with the Bavarian AI Tax Laboratory at the University of Technology Nuremberg and holds an academic background in economics from FAU, with research experience in the application of AI to tax law and taxation-related decision support.

To enable inter-rater reliability analysis while maintaining a feasible workload, 20 statements (19.8% of the total) were designated as overlap items and independently graded by all three evaluators. The remaining statements were distributed evenly among evaluators without duplication. Human grading followed the original examination rubric used in SteuerEx. For each statement, evaluators were provided with the original exam question, the full reference solution, and the LLM-generated answer with any reasoning traces removed. Evaluators assigned points on the original statement-specific scale, typically ranging from 0 to between 0.5 and 5.0 points, based on correctness, completeness, and legal reasoning quality. All evaluations were conducted independently without discussion or coordination.

During data processing, four DeepSeek-R1-671B cases were identified in which malformed model outputs prevented meaningful comparison with human judgments. These cases were excluded from the human-LLM agreement analysis but retained for inter-rater reliability calculations when all three human scores were available, as the human judgments themselves remained valid. Agreement among human evaluators and alignment between human judgments and the automated evaluator were quantified using standard reliability and rank-correlation statistics. The statistical estimators and uncertainty quantification procedures are described in the Statistical analysis section below. A representative example of the human grading interface and statement-level assessment process is shown in **Supplementary Figure 1**.

## Statistical analysis

All statistical analyses were performed using Python v3.11 with NumPy, SciPy, and statsmodels.

Prior to analysis, all human-assigned statement scores were normalized to percentages to account for differing maximum point values across statements. For a given statement $i$ and rater $j$, the normalized score $s_{ij}$ was computed as:

$$s_{ij} = \frac{\text{points}_{ij}}{\text{max\_points}_i} \times 100. \tag{17}$$

To assess alignment between human judgments and the automated LLM-based evaluator, rank correlation was measured using Kendall's $\tau_b$[65]. This statistic was chosen because it is robust to ties, does not assume normally distributed scores, and provides an interpretable measure of concordance for ordinal educational grading[66] data. For each statement $i$, the human reference score $\bar{s}_i$ was computed as the mean of all available human ratings,

$$\bar{s}_i = \frac{1}{m_i} \sum_{j=1}^{m_i} s_{ij}, \tag{18}$$

where $m_i \in \{1,2,3\}$ denotes the number of human evaluators who graded statement $i$. The corresponding automated score $a_i$ was obtained from the LLM-based evaluator and likewise normalized to a percentage scale. Kendall's $\tau_b$ between human and automated scores was computed as:

$$\tau_b = \frac{C - D}{\sqrt{(C + D + T_h)(C + D + T_a)}}, \tag{19}$$

Here, $C$ denotes the number of concordant pairs, $D$ the number of discordant pairs, $T_h$ the number of ties in human scores, and $T_a$ the number of ties in automated scores. Correlations were computed both for the pooled sample and separately for each evaluated model. Uncertainty in Kendall's $\tau_b$ was quantified using non-parametric bootstrapping with 10,000 resamples and replacement[67]. In each bootstrap iteration, statement-level paired score observations were resampled and $\tau_b$ recomputed. Ninety-five percent confidence intervals were derived from the 2.5th and 97.5th percentiles of the resulting bootstrap distribution. As a complementary measure of inter-rater reliability, we also computed pairwise Kendall's $\tau_b$ between human evaluators to quantify rank-order agreement.

The primary outcome measure for model benchmarking is the total score achieved on the SteuerEx benchmark, normalized to the maximum achievable score of 1,035.5 points and expressed as a percentage.

Uncertainty estimates for model performance were obtained using a non-parametric, points-constrained bootstrap with 1,000 resamples with replacement at the question level, reflecting the heterogeneous scoring structure of the benchmark. In each bootstrap iteration, questions were sampled with replacement until the sum of their maximum point values equaled the benchmark total, yielding a replicate that corresponds to a full benchmark with identical total available points. Bootstrap resampling was performed under a fixed randomization scheme to ensure exact reproducibility. For each model and replicate, earned points were summed over the sampled questions to obtain a bootstrapped total score, which was expressed as a percentage by dividing by 1,035.5 and multiplying by 100. From the resulting bootstrap distributions, we computed the mean, standard deviation, and 95% confidence intervals using the 2.5th and 97.5th percentiles. Statistical significance of differences between models was assessed using paired permutation tests on per-question point differences[68], with each model compared against SteuerLLM as the reference system under a paired[69] design. The test statistic was defined as the difference in overall percentage points, computed as the difference in total earned points divided by the total available points. A Monte Carlo null distribution was generated by randomly swapping model labels within each question, equivalently applying independent random sign flips to per-question point differences, repeated 10,000 times with a fixed random seed for reproducibility. Two-sided p-values were computed as the proportion of permuted differences at least as extreme as the observed difference, using a standard plus-one correction to avoid zero p-values. Resulting p-values were adjusted for multiple comparisons using the Benjamini-Hochberg false discovery rate procedure[70], with a significance threshold of 0.05.

# Data availability

The data supporting the findings of this study are mostly publicly available. The Public Data Instruct dataset used to train SteuerLLM was generated using the proposed Water Fountain algorithm and is derived exclusively from publicly accessible legal sources. This dataset is publicly released via a public Hugging Face repositor https://huggingface.co/datasets/windprak/steuerllm_instruct_dataset. In addition, the model was trained on a curated subset of publicly available web documents, matched by URL against the FineWeb corpus[71]. This dataset is likewise released via Hugging Face https://huggingface.co/datasets/windprak/steuerllm_pretraining_dataset.The SteuerEx benchmark is released via a public GitHub repository https://github.com/windprak/steuerllm_benchmark. The repository contains the benchmark specification, question material, and scripts used for model answer generation and score aggregation.

# Code availability

All source code, configurations, parameters, models, and workflows used in this study are publicly available.

The synthetic data generation workflows, including the Water Fountain Algorithm, as well as model fine-tuning configurations for SteuerLLM, are released via a separate GitHub repository https://github.com/windprak/steuerllm. The implementation is primarily developed in Python 3. Benchmark evaluation, scoring, and statistical analysis were conducted using the OpenAI Python SDK v1.77.0 (https://platform.openai.com) for model access and GPT-4o-based automated adjudication, together with pandas v2.2.3, NumPy v2.0.1, and SciPy v1.15.3 for data processing and statistical evaluation, including bootstrap resampling. HTTP-based endpoint communication was handled using requests v2.32.3. Figures were generated using matplotlib (v3.4+) and seaborn.

The synthetic data generation pipeline relies on aiohttp (v3.9.0+), aiofiles (v23.0.0+), pydantic (v2.0.0+), and PyYAML (v6.0.0+). Retrieval-based contextualization during data generation was implemented using the SearxNG metasearch engine (https://github.com/searxng/searxng), deployed via Docker for online web retrieval. Query generation additionally used a locally hosted Ollama runtime (v0.1+) serving llama3:8b-instruct-q8_0. Primary instruction-style question–answer generation was performed using Mistral-Large-Instruct-2407 (https://huggingface.co/mistralai/Mistral-Large-Instruct-2407).

Model fine-tuning was carried out using the Axolotl framework v0.10.0 (https://github.com/axolotl-ai-cloud/axolotl) with PyTorch v2.7.0+cu128 and Transformers v4.51.3 (https://github.com/huggingface/transformers). Additional components included PEFT v0.15.2, accelerate v1.6.0, DeepSpeed v0.15.4, bitsandbytes v0.45.4, flash-attn v2.7.4, and Weights & Biases (wandb) v0.19.11. Training was executed within the Axolotl container environment (axolotl.sif) using Apptainer to ensure reproducibility.

To mitigate risks of training data contamination and benchmark leakage, the LLM-based evaluation pipeline is not released publicly at this time. Instead, benchmark evaluation is performed via a dedicated external service https://steuerllm.i5.ai.fau.de/benchmark, which ensures controlled and consistent assessment while preventing direct access to evaluation prompts and grading logic. The complete benchmark, including the LLM-based evaluation components, will be released in full once the evaluation service is discontinued.

All locally deployed language models were sourced from Hugging Face, assessed and used between January and April 2025, and include the following models with their corresponding URLs:

- Qwen2.5-3B-Instruct: https://huggingface.co/Qwen/Qwen2.5-3B-Instruct
- Qwen2.5-7B-Instruct: https://huggingface.co/Qwen/Qwen2.5-7B-Instruct
- Qwen2.5-14B-Instruct: https://huggingface.co/Qwen/Qwen2.5-14B-Instruct
- Qwen2.5-32B-Instruct: https://huggingface.co/Qwen/Qwen2.5-32B-Instruct
- Qwen2.5-72B-Instruct: https://huggingface.co/Qwen/Qwen2.5-72B-Instruct
- Llama-3.2-3B-Instruct: https://huggingface.co/meta-llama/Llama-3.2-3B-Instruct
- Meta-Llama-3-8B-Instruct: https://huggingface.co/meta-llama/Meta-Llama-3-8B-Instruct
- DeepSeek-R1-Distill-Llama-70B: https://huggingface.co/deepseek-ai/DeepSeek-R1-Distill-Llama-70B
- DeepSeek-R1-671B: https://huggingface.co/deepseek-ai/DeepSeek-R1
- Ministral-8B-Instruct-2410: https://huggingface.co/mistralai/Ministral-8B-Instruct-2410
- Mistral-Small-Instruct-2409: https://huggingface.co/mistralai/Mistral-Small-Instruct-2409

- Gemma-3-4B-it: https://huggingface.co/google/gemma-3-4b-it
- Gemma-3-27B-it: https://huggingface.co/google/gemma-3-27b-it

All locally hosted models were served using vLLM v0.9.0 (https://github.com/vllm-project/vllm), with tensor parallelism set to the number of GPUs available per node. Models under 3 billion parameters were served without tensor parallelism. OpenAI-hosted models were accessed via direct REST API calls to the OpenAI endpoints (https://platform.openai.com). The exact OpenAI model versions used in this study were *gpt-4o-2024-08-06* and *gpt-4o-mini-2024-07-18*.

The domain-specific SteuerLLM models (Small-SteuerLLM and SteuerLLM) were developed at NHR@FAU. The Open-SteuerLLM (28B) model weights are publicly available via Hugging Face https://huggingface.co/windprak/open_steuerllm, together with an online demonstration interface https://steuerllm.i5.ai.fau.de/chat.

Most experiments, including benchmark inference and large-scale synthetic data generation, were conducted on GPU nodes equipped with NVIDIA H100 accelerators on NHR@FAU's Helma Cluster (https://doc.nhr.fau.de/clusters/helma/). Inference for extremely large-scale architectures requiring substantially higher memory capacity, including DeepSeek-R1-671B, was executed on nodes equipped with AMD Instinct MI300X accelerators. Additional computations, including early-stage data generation and endpoint consumption, were performed on a local workstation.

# Acknowledgements

This research is supported by BayernKI, the central infrastructure for the State of Bavaria to advance academic AI research. The authors gratefully acknowledge the HPC resources provided by the Erlangen National High Performance Computing Center (NHR@FAU) of the Friedrich-Alexander-Universität Erlangen-Nürnberg. NHR funding is provided by federal and Bavarian state authorities. NHR@FAU hardware is partially funded by the Deutsche Forschungsgemeinschaft (DFG) – 440719683.

# Author contributions

The formal analysis and study conceptualization were conducted by SW and STA. The original draft was written by SW, JS, LS, and STA and edited by STA. The experiments were performed by SW and LS. SW, LS, and JS developed the codes for analysis and pipeline; SW configured and maintained the LLM-serving infrastructure. The SteuerEx benchmark was curated by SW, LS, and QJ. The synthetic training data was curated by SW. The statistical analyses were performed by SW, JS, and STA. SW, JS, LS, FW, MM, HK, WG, AM, and STA provided technical expertise. SW, QJ, and SK provided financial expertise. All authors read the manuscript and agreed to the submission of this paper.

## Competing interests

SW is partially employed by DATEV eG, Germany. JS is partially employed by Siemens Healthineers AG, Germany. AM is an associate editor at *IEEE Transactions on Medical Imaging*. STA is an editorial board at Communications Medicine and *European Radiology Experimental*, and a trainee editorial board at *Radiology: Artificial Intelligence*. The other authors do not have any competing interests to disclose.

**References**

1. Brown, T. B. *et al.* Language models are few-shot learners. in *Proceedings of the 34th International Conference on Neural Information Processing Systems* vol. 159 1877–1901 (2020).
2. Vaswani, A. *et al.* Attention Is All You Need. in *NIPS'17: Proceedings of the 31st International Conference on Neural Information Processing Systems* 6000–6010 (2017).
3. Wind, S. *et al.* Multi-step retrieval and reasoning improves radiology question answering with large language models. *npj Digit. Med.* **8**, 790 (2025).
4. Wei, J. *et al.* Chain-of-thought prompting elicits reasoning in large language models. *Advances in neural information processing systems* **35**, 24824–24837 (2022).
5. Blair-Stanek, A., Holzenberger, N. & Van Durme, B. Can gpt-3 perform statutory reasoning? in *Proceedings of the Nineteenth International Conference on Artificial Intelligence and Law* 22–31 (2023).
6. Dugac, G. & Altwicker, T. Classifying legal interpretations using large language models. *Artificial Intelligence and Law* 1–19 (2025).
7. Dietterich, T. G. Steps toward robust artificial intelligence. *Ai Magazine* **38**, 3–24 (2017).
8. Wei, J. *et al.* Finetuned language models are zero-shot learners. *arXiv preprint arXiv:2109.01652* (2021).
9. Marcus, G. The next decade in AI: four steps towards robust artificial intelligence. *arXiv preprint arXiv:2002.06177* (2020).
10. Lewis, P. *et al.* Retrieval-Augmented Generation for Knowledge-Intensive NLP Tasks. in *Advances in Neural Information Processing Systems* (eds Larochelle, H., Ranzato, M., Hadsell, R., Balcan, M. F. & Lin, H.) vol. 33 9459–9474 (Curran Associates, Inc., 2020).
11. Chalkidis, I. *et al.* LexGLUE: A benchmark dataset for legal language understanding in English. in *Proceedings of the 60th Annual Meeting of the Association for Computational Linguistics (Volume 1: Long Papers)* 4310–4330 (2022).
12. Chalkidis, I., Fergadiotis, M., Malakasiotis, P., Aletras, N. & Androutsopoulos, I. LEGAL-BERT: The muppets straight out of law school. *arXiv preprint arXiv:2010.02559* (2020).
13. Li, J., Bhambhoria, R. & Zhu, X. Parameter-efficient legal domain adaptation. *arXiv preprint arXiv:2210.13712* (2022).
14. Colombo, P. *et al.* Saullm-54b & saullm-141b: Scaling up domain adaptation for the legal domain. *Advances in Neural Information Processing Systems* **37**, 129672–129695 (2024).
15. Hendrycks, D. *et al.* Measuring massive multitask language understanding. *arXiv preprint arXiv:2009.03300* (2020).
16. Rabelo, J. *et al.* Overview and discussion of the competition on legal information extraction/entailment (COLIEE) 2021. *The Review of Socionetwork Strategies* **16**, 111–133 (2022).
17. Bernet, H. & Berteloot, P. EUR-Lex: A multilingual on-line website for European Union law. *International Review of Law Computers & Technology* **20**, 337–339 (2006).
18. Aumiller, D., Chouhan, A. & Gertz, M. EUR-lex-sum: A multi-and cross-lingual dataset for long-form summarization in the legal domain. *arXiv preprint arXiv:2210.13448* (2022).
19. Kirchhof, P. *Einkommensteuergesetz: Kommentar*. (Verlag Dr. Otto Schmidt, 2012).
20. Lang, J. & Tipke, K. *Steuerrecht*. (Otto Schmidt, 2020).
21. Grechenig, K. & Gelter, M. The transatlantic divergence in legal thought: American law and economics vs. German doctrinalism. *Hastings Int'l & Comp. L. Rev.* **31**, 295 (2008).
22. McGinnis, J. O. & Pearce, R. G. The great disruption: How machine intelligence will transform the role of lawyers in the delivery of legal services. *Fordham L. Rev.* **82**, 3041 (2013).
23. Wang, A. *et al.* GLUE: A Multi-Task Benchmark and Analysis Platform for Natural Language Understanding. in *Proceedings of the 2018 EMNLP Workshop BlackboxNLP: Analyzing and*

*Interpreting Neural Networks for NLP* 353–355 (Association for Computational Linguistics, Brussels, Belgium, 2018). doi:10.18653/v1/W18-5446.
24. Son, G. *et al.* KMMLU: Measuring Massive Multitask Language Understanding in Korean. in *Proceedings of the 2025 Conference of the Nations of the Americas Chapter of the Association for Computational Linguistics: Human Language Technologies (Volume 1: Long Papers)* 4076–4104 (Association for Computational Linguistics, Albuquerque, New Mexico, 2025). doi:10.18653/v1/2025.naacl-long.206.
25. Rajpurkar, P., Zhang, J., Lopyrev, K. & Liang, P. SQuAD: 100,000+ Questions for Machine Comprehension of Text. in *Proceedings of the 2016 Conference on Empirical Methods in Natural Language Processing* 2383–2392 (Association for Computational Linguistics, Austin, Texas, 2016). doi:10.18653/v1/D16-1264.
26. Wolfson, T. *et al.* Break It Down: A Question Understanding Benchmark. *Transactions of the Association for Computational Linguistics* **8**, 183–198 (2020).
27. Kaplan, J. *et al.* Scaling laws for neural language models. *arXiv preprint arXiv:2001.08361* (2020).
28. Hoffmann, J. *et al.* Training compute-optimal large language models. *arXiv preprint arXiv:2203.15556* (2022).
29. Zheng, L. *et al.* Judging llm-as-a-judge with mt-bench and chatbot arena. *Advances in neural information processing systems* **36**, 46595–46623 (2023).
30. Schick, T. & Schütze, H. Generating datasets with pretrained language models. *arXiv preprint arXiv:2104.07540* (2021).
31. Eldan, R. & Li, Y. Tinystories: How small can language models be and still speak coherent english? *arXiv preprint arXiv:2305.07759* (2023).
32. Katz, D. M., Bommarito II, M. J. & Blackman, J. A general approach for predicting the behavior of the Supreme Court of the United States. *PloS one* **12**, e0174698 (2017).
33. Katz, D. M. & Bommarito, M. J. Measuring the complexity of the law: the United States Code. *Artificial intelligence and law* **22**, 337–374 (2014).
34. Surden, H. Machine learning and law. *Wash. L. Rev.* **89**, 87 (2014).
35. Guu, K., Lee, K., Tung, Z., Pasupat, P. & Chang, M. Retrieval augmented language model pre-training. in *International conference on machine learning* 3929–3938 (PMLR, 2020).
36. Izacard, G. & Grave, E. Leveraging passage retrieval with generative models for open domain question answering. in *Proceedings of the 16th conference of the european chapter of the association for computational linguistics: main volume* 874–880 (2021).
37. Crocetti, G. Textual Spatial Cosine Similarity. Preprint at https://doi.org/10.48550/arXiv.1505.03934 (2015).
38. Shao, M., Basit, A., Karri, R. & Shafique, M. Survey of different Large Language Model Architectures: Trends, Benchmarks, and Challenges. *IEEE Access* **12**, 188664–188706 (2024).
39. Huang, G., Sun, Y., Liu, Z., Sedra, D. & Weinberger, K. Q. Deep Networks with Stochastic Depth. in *Computer Vision – ECCV 2016* (eds Leibe, B., Matas, J., Sebe, N. & Welling, M.) vol. 9908 646–661 (Springer International Publishing, Cham, 2016).
40. He, K., Zhang, X., Ren, S. & Sun, J. Identity Mappings in Deep Residual Networks. in *Computer Vision – ECCV 2016* (eds Leibe, B., Matas, J., Sebe, N. & Welling, M.) vol. 9908 630–645 (Springer International Publishing, Cham, 2016).
41. Howard, J. & Ruder, S. Universal language model fine-tuning for text classification. *arXiv preprint arXiv:1801.06146* (2018).
42. Kirkpatrick, J. *et al.* Overcoming catastrophic forgetting in neural networks. *Proc. Natl. Acad. Sci. U.S.A.* **114**, 3521–3526 (2017).
43. Bengio, Y., Ducharme, R., Vincent, P. & Jauvin, C. A neural probabilistic language model. *The Journal of Machine Learning Research* **3**, 1137–1155.
44. Loshchilov, I. & Hutter, F. Decoupled Weight Decay Regularization. in *Proceedings of Proceedings of Seventh International Conference on Learning Representations (ICLR) 2019* (New Orleans, LA, USA, 2019).

45. Shoeybi, M. *et al.* Megatron-LM: Training Multi-Billion Parameter Language Models Using Model Parallelism. Preprint at https://doi.org/10.48550/arXiv.1909.08053 (2020).
46. Narayanan, D. *et al.* Efficient large-scale language model training on GPU clusters using megatron-LM. in *Proceedings of the International Conference for High Performance Computing, Networking, Storage and Analysis* 1–15 (ACM, St. Louis Missouri, 2021). doi:10.1145/3458817.3476209.
47. Schnitger, A. & Fehrenbacher, O. *Kommentar Körperschaftsteuer KStG*. (Springer, 2012).
48. Lloyd, A. C. The Logical Form of Law Statements. *Mind* **64**, 312–318 (1955).
49. Zhong, H. *et al.* How Does NLP Benefit Legal System: A Summary of Legal Artificial Intelligence. in *Proceedings of the 58th Annual Meeting of the Association for Computational Linguistics* 5218–5230 (Association for Computational Linguistics, Online, 2020). doi:10.18653/v1/2020.acl-main.466.
50. Holtzman, A., Buys, J., Du, L., Forbes, M. & Choi, Y. The Curious Case of Neural Text Degeneration. in *ICLR 2020* (2020).
51. Touvron, H. *et al.* LLaMA: Open and Efficient Foundation Language Models. Preprint at http://arxiv.org/abs/2302.13971 (2023).
52. Grattafiori, A. *et al.* The Llama 3 Herd of Models. Preprint at https://doi.org/10.48550/arXiv.2407.21783 (2024).
53. Bai, J. *et al.* Qwen technical report. *arXiv preprint arXiv:2309.16609* (2023).
54. DeepSeek-AI *et al.* DeepSeek-R1: Incentivizing Reasoning Capability in LLMs via Reinforcement Learning. Preprint at https://doi.org/10.48550/arXiv.2501.12948 (2025).
55. Team, G. *et al.* Gemma: Open models based on gemini research and technology. *arXiv preprint arXiv:2403.08295* (2024).
56. Team, G. *et al.* Gemma 3 technical report. *arXiv preprint arXiv:2503.19786* (2025).
57. Kojima, T., Gu, S. (Shane), Reid, M., Matsuo, Y. & Iwasawa, Y. Large Language Models are Zero-Shot Reasoners. in *Advances in Neural Information Processing Systems* (eds Koyejo, S. et al.) vol. 35 22199–22213 (Curran Associates, Inc., 2022).
58. OpenAI. GPT-4 Technical Report. Preprint at http://arxiv.org/abs/2303.08774 (2023).
59. Zhao, W., Strube, M. & Eger, S. DiscoScore: Evaluating Text Generation with BERT and Discourse Coherence. in *Proceedings of the 17th Conference of the European Chapter of the Association for Computational Linguistics* 3865–3883 (Association for Computational Linguistics, Dubrovnik, Croatia, 2023). doi:10.18653/v1/2023.eacl-main.278.
60. Raji, I. D. *et al.* Closing the AI accountability gap: defining an end-to-end framework for internal algorithmic auditing. in *Proceedings of the 2020 Conference on Fairness, Accountability, and Transparency* 33–44 (ACM, Barcelona Spain, 2020). doi:10.1145/3351095.3372873.
61. Hendrycks, D., Carlini, N., Schulman, J. & Steinhardt, J. Unsolved Problems in ML Safety. Preprint at https://doi.org/10.48550/arXiv.2109.13916 (2022).
62. Bommarito, M. & Katz, D. M. GPT Takes the Bar Exam. Preprint at https://doi.org/10.48550/arXiv.2212.14402 (2022).
63. William G., C. *Sampling Techniques*. (Wiley, 1991).
64. Shrout, P. E. & Fleiss, J. L. Intraclass correlations: Uses in assessing rater reliability. *Psychological Bulletin* **86**, 420–428 (1979).
65. Kendall, M. G. A New Measure of Rank Correlation. *Biometrika* **30**, 81 (1938).
66. Cohen, J. Weighted kappa: Nominal scale agreement provision for scaled disagreement or partial credit. *Psychological Bulletin* **70**, 213–220 (1968).
67. Efron, B. & Tibshirani, R. J. *An Introduction to the Bootstrap*. (Chapman and Hall/CRC, 1994). doi:10.1201/9780429246593.
68. Konietschke, F. & Pauly, M. Bootstrapping and permuting paired t-test type statistics. *Stat Comput* **24**, 283–296 (2014).
69. Dieterich, T. G. Approximate Statistical Tests for Comparing Supervised Classification Learning Algorithms. *Neural Computation* **10**, 1895–1923 (1998).

70. Benjamini, Y. & Hochberg, Y. Controlling the False Discovery Rate: A Practical and Powerful Approach to Multiple Testing. *Journal of the Royal Statistical Society Series B: Statistical Methodology* **57**, 289–300 (1995).
71. Penedo, G. *et al.* The fineweb datasets: Decanting the web for the finest text data at scale. *Advances in Neural Information Processing Systems* **37**, 30811–30849 (2024).

# Supplementary information

**Supplementary Figure 1:** Web-based human evaluation interface used for statement-level grading of LLM-generated tax law answers. The interface presents the original exam question, the complete reference solution, and the LLM response with highlighted content, followed by an individual examiner statement to be graded. Human evaluators assign partial or full credit according to the original examination rubric, with automatic logging of scores for subsequent inter-rater reliability and human-LLM agreement analyses.

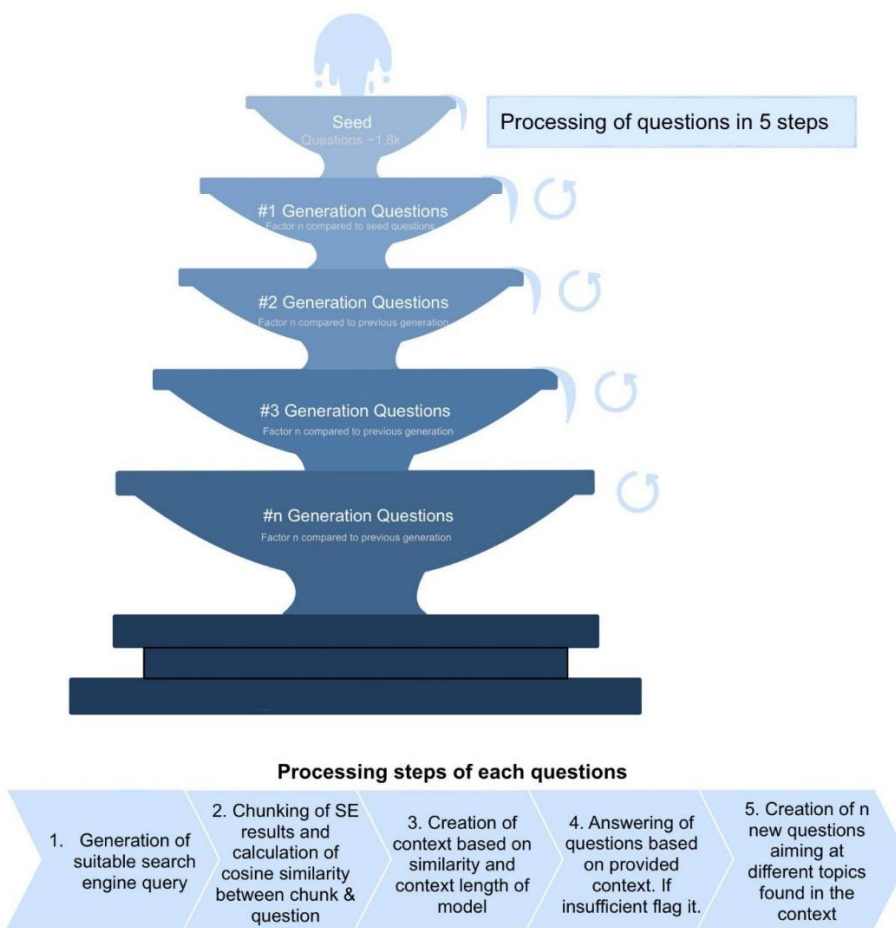

**Supplementary Figure 2:** Iterative structure of the Water Fountain Algorithm for synthetic data generation. The diagram illustrates the multi-stage pipeline used to generate synthetic question-answer pairs for German tax law. Starting from approximately 1,800 authentic examination questions, each generation cycle applies five steps: query generation, retrieval and chunking with similarity scoring, context construction under a fixed context length, answer generation with explicit insufficiency flagging, and creation of multiple new thematically diversified questions. Each layer corresponds to one generation, leading to progressive dataset expansion and increasing topical diversity across iterations.

**Supplementary Table 1:** Subtopic-level composition of the SteuerEx benchmark within selected tax law categories. The table reports the distribution of questions and maximum achievable points across statutory subtopics within individual examinations. Subtopics reflect dominant doctrinal areas covered by each exam and are shown to illustrate internal heterogeneity within benchmark categories. Each question is assigned to exactly one subtopic and counted once. This breakdown is provided for transparency and to contextualize differences in point weighting and difficulty across categories; it is not used directly in model evaluation or statistical analysis.

| Category | Exam name | Subtopic (official German name) | Subtopic (English translation) | Total questions [n] | Maximum achievable points |
|---|---|---|---|---|---|
| Corporate tax (Unternehmensbesteuerung) | UnternehmenSt SS18 | Gewerbesteuer | Trade tax | 1 | 12.0 |
| | | Körperschaftsteuer | Corporate income tax | 4 | 36.0 |
| | | Umwandlungssteuer | Reorganization tax | 2 | 6.0 |
| | UnternehmenSt SS19 | Gewerbesteuer | Trade tax | 1 | 13.0 |
| | | Körperschaftsteuer | Corporate income tax | 6 | 34.0 |
| | | Umwandlungssteuer | Reorganization tax | 1 | 6.0 |
| | UnternehmenSt SS20 | Gewerbesteuer | Trade tax | 1 | 19.0 |
| | | Körperschaftsteuer | Corporate income tax | 5 | 30.0 |
| | | Umwandlungssteuer | Reorganization tax | 1 | 7.0 |
| | UnternehmenSt SS22 | Gewerbesteuer | Trade tax | 1 | 10.5 |
| | | Körperschaftsteuer | Corporate income tax | 10 | 36.0 |
| | | Umwandlungssteuer | Reorganization tax | 3 | 6.0 |
| | UnternehmenSt SS23 | Körperschaftsteuer | Corporate income tax | 4 | 20.0 |
| | | Körperschaft- & Gewerbesteuer | Corporate & trade tax | 3 | 24.0 |
| | | Umwandlungssteuer | Reorganization tax | 1 | 2.0 |
| Fundamentals of tax law (Grundlagen des Steuerrechts) | GrldStR WS23/24 | International tax law | International tax law | 1 | 7.0 |
| | | Umsatzsteuer | Value-added tax | 4 | 20.0 |

**Supplementary Table 2:** Exam-level composition of SteuerEx and distribution of student examination performance across individual tax law exams. Each exam-semester pair corresponds to one examination included in the SteuerEx category mapping. For each exam, the table reports the number of students (where available), the number of benchmark questions assigned to the category, and the maximum achievable benchmark points contributed by that exam (as used in the SteuerEx evaluation). Student performance is reported as lowest, average, and highest raw scores under the original grading scheme. Categories are mutually exclusive and each benchmark question is assigned to exactly one category. Student performance statistics are unavailable for some benchmark exams (marked N/A) because the student outcome dataset does not contain exam-level score distributions for those specific exam-semester pairs.

| Category | Exam name | Total students [n] | Total questions [n] | Maximum achievable points | Lowest grade (raw points) | Average grade (raw points) | Highest grade (raw points) |
|---|---|---|---|---|---|---|---|
| Corporate tax (Unternehmensbesteuerung) | UnternehmenSt SS18 | 80 | 7 | 54.0 | 15.0 | 34.2 | 45.5 |
| | UnternehmenSt SS19 | 68 | 8 | 53.0 | 14.5 | 31.1 | 44.5 |
| | UnternehmenSt SS20 | 84 | 7 | 56.0 | 8.5 | 30.6 | 44.0 |
| | UnternehmenSt SS22 | 86 | 14 | 52.5 | 9.5 | 29.6 | 51.5 |
| | UnternehmenSt SS23 | 89 | 8 | 46.0 | 0.5 | 30.5 | 62.0 |
| Fiscal code (Abgabenordnung) | AO SS20 | 16 | 2 | 62.0 | 15.5 | 35.6 | 50.0 |
| | AO WS16/17 | 29 | 1 | 67.0 | 7.0 | 30.1 | 42.5 |
| Fundamentals of tax law (Grundlagen des Steuerrechts) | GrldStR SS21 | 10 | 12 | 57.0 | 5.5 | 23.9 | 37.0 |
| | GrldStR WS19/20 | 130 | 15 | 58.0 | 6.0 | 29.8 | 52.0 |
| | GrldStR WS21/22 | N/A | 6 | 50.0 | N/A | N/A | N/A |
| | GrldStR WS22/23 | 72 | 11 | 49.0 | 9.5 | 29.1 | 45.5 |
| | GrldStR WS23/24 | 61 | 12 | 55.0 | 11.0 | 25.7 | 44.0 |
| Income tax (Einkommensteuerrecht) | EStR WS19/20 | N/A | 1 | 63.0 | N/A | N/A | N/A |
| | EStR WS20/21 | 27 | 1 | 66.0 | 34.5 | 46.3 | 57.0 |
| | EStR WS21/22 | 32 | 2 | 60.0 | 15.5 | 34.5 | 44.5 |
| Taxation of partnerships (Besteuerung von Personengesellschaften) | PersG SS19 | 16 | 4 | 66.0 | 30.5 | 39.7 | 58.0 |
| Value-added tax (Umsatzsteuerrecht) | USt SS21 | 31 | 2 | 50.0 | 16.0 | 30.0 | 45.5 |
| | USt SS22 | 22 | 2 | 71.0 | 12.5 | 35.5 | 57.5 |

**Supplementary Table 3:** Human evaluation and validation of automated statement-level grading. Summary of the human evaluation study used to assess both inter-rater reliability among human graders and agreement between human judgments and the automated LLM-based evaluator. The table reports sample composition, exclusion counts due to malformed model outputs, and overlap size used for reliability analysis. Human-human agreement is quantified on the overlap subset using the intraclass correlation coefficient (ICC(2,1)) for absolute agreement, together with descriptive agreement statistics and pairwise Spearman rank correlations between evaluators. Human-LLM agreement is reported using Kendall's τ rank correlation between averaged human scores and automated evaluator scores, both overall and stratified[1] by model. Confidence intervals denote 95% bootstrap intervals. N/A indicates metrics not applicable at the model level.

| Metric | Overall | DeepSeek-R1-671B | Llama-3.2-3B-it | SteuerLLM | GPT-4o-mini |
|---|---|---|---|---|---|
| **Sample composition (n = 59)** | | | | | |
| Unique statement-level items [n] | 101 | 17 | 15 | 15 | 16 |
| Items with valid human + LLM score [n] | 59 | 13 | 15 | 15 | 16 |
| Items excluded due to malformed model output [n] | 4 | 4 | 0 | 0 | 0 |
| Overlap items graded by all 3 raters [n] | 20 | N/A | N/A | N/A | N/A |
| **Inter-rater reliability (human vs. human; overlap set only n = 20)** | | | | | |
| ICC(2,1) (absolute agreement) | 0.367 | N/A | N/A | N/A | N/A |
| Perfect agreement among all 3 raters [n (%)] | 3 (15.0%) | N/A | N/A | N/A | N/A |
| Mean score range across raters [%] | 37.99 | N/A | N/A | N/A | N/A |
| Mean within-item SD across raters [%] | 21.07 | N/A | N/A | N/A | N/A |
| Mean absolute difference across raters [%] | 15.84 | N/A | N/A | N/A | N/A |
| Spearman ρ (HIWI_1 vs. HIWI_2) | 0.698 (P=0.0006) | N/A | N/A | N/A | N/A |
| Spearman ρ (HIWI_1 vs. HIWI_3) | 0.234 (P=0.310) | N/A | N/A | N/A | N/A |
| Spearman ρ (HIWI_2 vs. HIWI_3) | 0.373 (P=0.105) | N/A | N/A | N/A | N/A |
| **Human vs. automated evaluator agreement (Kendall's τ)** | | | | | |
| Kendall's τ | 0.718 | 0.756 | 0.730 | 0.740 | 0.732 |
| 95% CI for τ | [0.599, 0.820] | [0.460, 0.954] | [0.380, 0.957] | [0.455, 0.935] | [0.460, 0.900] |

**Supplementary Table 4:** Overview of data cleaning and filtering steps applied to the synthetic dataset. This table summarizes the exclusion categories applied during the preprocessing of the raw synthetic data. The cleaning process involved removing exact duplicates, excluding flagged answers based on error indicators, and applying context-based filtering to ensure data quality. A total of 120,625 question-answer pairs were removed to improve dataset diversity, validity, and consistency. N/A: not available.

| Category | Instances | Description |
| --- | --- | --- |
| **Initial data cleansing** | | |
| Exact duplicates removed | 47,555 | Identified and eliminated duplicates and seed questions to avoid redundancy and ensure dataset diversity. |
| **Exact flag-based exclusion** | | |
| Flagged answers (exact match) | 11,483 | Answers consisting solely of the flag string (e.g., "Es tut mir Leid") indicating an error or lack of sufficient context. |
| **Context-based filtering** | | |
| Total excluded (context-based filtering) | 61,587 | Combined exclusions due to flagged partial matches and insufficient retrieved sources. |
| Flagged answers (partial match) | N/A | Responses containing variations of the flag string within the text, signaling insufficient context or improper formatting. |
| Insufficient retrieved sources | N/A | Tuples where fewer than 3 relevant sources were retrieved from SearXNG, leading to likely insufficient context for a valid response. |

**Supplementary Table 5:** Composition and categorization of the final synthetic training dataset. The final dataset used for fine-tuning SteuerLLM consists of 485,092 question-answer pairs, divided into two main groups: Primary Synthetic Generation and Context-Supported Generation. These categories encompass multiple subtypes targeting specific tax domains and task types, including standard advisory questions, article-based prompts, income tax, accounting, and comment-style analytical tasks. This structure should enable targeted model training for domain adaptation and robust generalization across a broad range of tax-related use cases.

| Category | Instances | Description |
| --- | --- | --- |
| **Primary synthetic generation** | | |
| Standard data | 215,391 | Base dataset covering various tax law issues derived systematically from a tax advisory practitioner. |
| Diversity data | 62,289 | Designed to increase diversity, featuring alternative task types and complex scenarios to improve generalization. |
| Income tax data | 7,286 | Focused on income tax law, covering e.g. the seven types of income (EStG), taxation of self-employment, and deductibility. |
| Article questions | 20,186 | Questions generated on specific legal paragraphs to improve the model's ability to reference and interpret them. |
| Accounting record questions | 2,700 | Targeting accounting processes, generated using GPT-4o for better performance in financial accounting. |
| Comment questions | 4,378 | Commentary-style questions that analyze and scrutinize tax law articles. |
| **Context-supported generation** | | |
| Chunk questions | 172,862 | Generated based on retrieved tax law contexts, ensuring questions relate directly to the given context. |

**Supplementary Table 6:** Comparison of observed points-weighted accuracy and points-constrained bootstrap mean accuracy across evaluated models. The table reports, for each model, the observed overall accuracy on the SteuerEx benchmark, the accuracy obtained from 1,000 points-constrained bootstrap replicates constructed to match the fixed total point budget of 1,035.5 points, and the resulting shift in percentage points. This comparison quantifies the extent to which the constrained resampling design alters model-level performance estimates relative to its observed benchmark outcome.

| Model | Observed accuracy [%] | Bootstrap mean accuracy [%] | Shift (pp) |
| --- | --- | --- | --- |
| Open-SteuerLLM | 29.91 | 23.23 | −6.68 |
| DeepSeek-R1-671B | 44.19 | 38.54 | −5.65 |
| GPT-4o-mini | 26.90 | 21.81 | −5.09 |
| SteuerLLM | 32.93 | 28.38 | −4.55 |
| DeepSeek-R1-Distill-Llama-70B | 23.73 | 19.75 | −3.98 |
| Small-SteuerLLM | 20.11 | 16.49 | −3.62 |
| Gemma-3-27B-it | 26.41 | 22.80 | −3.61 |
| Meta-Llama-3-8B-it | 10.76 | 7.63 | −3.13 |
| Mistral-Small-it-2409 | 22.69 | 20.49 | −2.20 |
| Qwen2.5-72B-it | 20.69 | 18.89 | −1.80 |
| Gemma-3-4B-it | 12.63 | 10.91 | −1.72 |
| Ministral-8B-it-2410 | 14.83 | 13.14 | −1.69 |
| Qwen2.5-14B-it | 13.98 | 12.48 | −1.50 |
| Qwen2.5-3B-it | 6.11 | 5.21 | −0.90 |
| Qwen2.5-32B-it | 18.40 | 17.94 | −0.46 |
| Llama-3.2-3B-it | 8.28 | 8.70 | +0.42 |
| Qwen2.5-7B-it | 11.01 | 11.61 | +0.60 |

# Supplementary Note 1: Points-constrained bootstrap and permutation-based inference for the SteuerEx benchmark

The SteuerEx benchmark evaluates large language models (LLMs) on German tax law examination questions that carry heterogeneous maximum point values, ranging from 1 to 67 points per question across 115 questions totaling 1,035.5 available points. The primary performance metric is the points-weighted accuracy: the ratio of total earned points to total available points, expressed as a percentage. This metric reflects the examination conventions under which German tax law proficiency is assessed, where questions of greater complexity carry proportionally greater weight.

Quantifying uncertainty around this metric requires a resampling procedure that respects the heterogeneous point structure of the benchmark. We need two types of analysis. First, we require descriptive uncertainty intervals for each model's performance, characterizing the range of accuracy values that would be expected under variation in the composition of the examination. Second, we require formal hypothesis tests to determine whether observed pairwise differences between models are statistically significant. We address these tasks with complementary methods: a points-constrained bootstrap for the former and paired permutation tests for the latter. In a standard non-parametric bootstrap, resampling a fixed number of observations is appropriate because each observation contributes equally to the statistic of interest. In our setting, however, each question carries a distinct maximum point. A standard bootstrap would resample a fixed number of questions and consequently produce replicate datasets with varying total point denominators.

To address this, we implement a points-constrained bootstrap[2]. Each of $B = 1{,}000$ bootstrap replicates is constructed by drawing questions with replacement from the full question pool until the cumulative sum of maximum points equals exactly the target budget $T = 1{,}035.5$. The algorithm proceeds sequentially: at each step, a question is sampled at random from among those whose maximum point value does not exceed the remaining budget, and the budget is decremented accordingly. When the remaining budget reaches zero, the replicate is complete. Situations in which no feasible draw remains are rare; if encountered, the replicate is restarted. For each replicate b, the points-weighted accuracy is computed as the sum of earned points divided by $T$. The bootstrap distribution of these 1,000 accuracy values is then summarized by its mean, standard deviation, and 2.5th and 97.5th percentiles to form a 95% percentile confidence interval.

For pairwise significance testing, we use paired[3] permutation tests. For each question, we compute the score difference between two models. Under the null hypothesis of no performance difference, the sign of this difference is arbitrary. We generate 10,000 random sign-flip permutations and compute the permuted test statistic for each. The two-tailed p-value is the proportion of permutations yielding a test statistic at least as extreme as the observed value, with a continuity correction of +1 in both numerator and denominator. All p-values are subsequently adjusted for multiple comparisons using the Benjamini-Hochberg procedure at a false discovery rate[4] of $\alpha = 0.05$.

These two procedures serve different inferential purposes. The bootstrap summarizes variability under a fixed-point budget resampling scheme, whereas the permutation test quantifies evidence for a difference on the observed paired question outcomes.

The constrained bootstrap has a direct consequence for question inclusion probabilities. In a single replicate, the number of times a 1-point question can be drawn is bounded only by the total budget, up to 1,035 times in principle, whereas a 67-point question can appear at most 15 times before exhausting the budget. Moreover, the sequential nature of the sampling procedure introduces path dependence: early selection of high-weight items reduces the allowable choices later in the replicate, increasing the relative likelihood that low-weight questions appear in the final steps.

This shift in inclusion probabilities results in a systematic difference between the bootstrap mean and the observed accuracy. Empirically, most models show a downward shift of the constrained bootstrap mean relative to the observed score, with shifts ranging from −6.7 to +0.6 percentage points. The size of this shift varies by model and can influence pairwise bootstrap differences when two models have different sensitivities to question weight. For example, models such as Qwen2.5-32B-it[5], Llama-3.2-3B-it[6,7] and Qwen2.5-3B-it[5] show bootstrap means that are very close to their observed performance. Larger shifts appear in models such as Open-SteuerLLM, DeepSeek-R1-671B[8], GPT-4o-mini, and SteuerLLM. A key limitation becomes visible when comparing models with different magnitudes of shift. For example, Open-SteuerLLM shows a larger downward shift than SteuerLLM. Because the shifts differ across models, the gap between their bootstrap means does not always follow the same pattern as the observed difference used in permutation testing.

For these reasons, we interpret constrained-bootstrap intervals as descriptive uncertainty summaries under a fixed-point budget design, and we rely on permutation tests for formal inference about pairwise differences between models. **Supplementary Table 6** reports the observed and constrained-bootstrap means for all models to make the magnitude and direction of these shifts transparent.

# Supplementary Note 2: Tax-law question type taxonomy used for synthetic data generation

To enable systematic and scalable generation of high-quality synthetic training data, we defined a taxonomy of tax-law question types that reflect recurring reasoning patterns in German tax examinations and professional tax practice. The taxonomy was developed in collaboration with professional tax advisors and served as a structural guide for prompt design and automated question generation within the Water Fountain Algorithm. The question types are not used as explicit supervision labels during training or evaluation but ensure broad coverage of doctrinal, procedural, and computational reasoning tasks.

The final taxonomy comprises 18 question types:

1. **Classification tasks**
   Assignment of facts, income, or transactions to legally defined tax categories or income types under statutory provisions.

2. **Fill-in-the-blank tasks**
   Completion of missing legal terms, statutory references, or numerical thresholds in tax-law statements or provisions.

3. **Sequence tasks**
   Ordering of procedural steps in tax assessments, filings, audits, or administrative processes.

4. **Text comprehension tasks**
   Interpretation of statutory provisions, administrative guidance, or case law excerpts and explanation of their legal implications.

5. **Argumentation tasks**
   Justification of tax-law positions through structured legal reasoning and reference to applicable norms.

6. **Matching tasks**
   Mapping of tax concepts, income types, or legal consequences to predefined categories or classifications.

7. **Extension tasks**
   Completion of incomplete legal enumerations or factual descriptions based on statutory structure.

8. **Explanatory tasks**
   Structured explanation of tax-law concepts or mechanisms in coherent natural language.

9. **Rule extraction tasks**
   Identification of applicable statutory requirements or conditions from legal texts.

10. **Comparison tasks**
    Systematic comparison of alternative tax treatments, income types, or legal regimes.

11. **Question generation tasks**
    Formulation of new tax-law questions based on a given statutory provision or legal concept.

12. **Translation tasks**
    Reformulation of complex tax-law provisions into simplified or non-technical language.

13. **Comprehension and commentary tasks**
    Analytical discussion of statutory provisions, including contextual interpretation and illustrative examples.

14. **Accounting and booking tasks**
    Derivation of accounting entries and identification of affected accounts in tax-relevant business transactions.

15. **Framework analysis tasks**
    Identification of relevant tax-law and legal conditions governing a broader factual constellation.

16. **Correction tasks**
    Detection and correction of incorrect tax treatments, calculations, or legal assumptions.

17. **Source identification tasks**
    Identification of relevant statutory provisions, administrative guidance, or legal sources applicable to a given scenario.

18. **Supplementary optimization tasks**
    Identification of additional tax planning options or legally permissible optimizations beyond an existing solution.

Representative examples of question-answer pairs generated under these categories are shown in **Supplementary Table 1**. Together, this taxonomy ensures that the synthetic dataset covers a wide spectrum of reasoning demands encountered in German tax law, ranging from statutory interpretation and procedural reasoning to numerical computation and accounting integration.

# Supplementary Note 3: Representative examples of synthetic question-answer pairs generated by the Water Fountain Algorithm

To illustrate the diversity, legal grounding, and structural complexity of the synthetic training data produced by the Water Fountain Algorithm, we provide representative examples of generated question-answer pairs across multiple generation categories. These examples are not intended to be exhaustive or statistically representative, but to qualitatively demonstrate how different prompt configurations, contextual retrieval strategies, and task types are reflected in the resulting data. All examples are drawn from the validated synthetic dataset used to train SteuerLLM.

**Chunk-based questions**

This category comprises questions generated from bounded textual contexts assembled via semantic retrieval from authoritative tax-law sources. The example concerns the tax treatment of a leasing special payment (Leasingsonderzahlung) for a business vehicle under the income surplus accounting method. The generated answer demonstrates period allocation over the contract term, explicit reference to relevant case law of the Bundesfinanzhof, numerical computation, and integration with the 1% rule cost cap. This category emphasizes grounded reasoning based on retrieved legal context and realistic numerical detail.

**Income tax questions**

Income tax questions focus on scenario-based application of statutory provisions, often involving valuation, classification, and interaction between different tax regimes. The illustrated example addresses the transfer of a business vehicle into the private assets of a managing director and discusses valuation under §6 Abs. 1 Nr. 4 Satz 2 EStG, alternative assessment methods, and the concept of hidden profit distributions. The generated answer combines doctrinal explanation, statutory citation, and illustrative numerical examples.

**Standard advisory scenarios**

This category reflects broadly framed advisory-style questions typical of professional tax consulting. The example discusses the outsourcing of an IT department into a separate GmbH and analyzes the resulting implications for value-added tax and trade tax. The answer integrates multiple statutory frameworks, including UStG and GewStG, and demonstrates structured legal analysis across corporate boundaries.

**Commentary and interpretation questions**

Commentary questions require interpretive discussion of individual statutory provisions. The example focuses on §382 AO and explains its function as a blanket provision in customs-related administrative offenses. The generated answer contextualizes the norm, outlines its legal scope, and provides practical examples illustrating its application in enforcement scenarios.

**Article-based questions.**

These questions center on detailed interpretation of specific statutory articles, often involving threshold calculations and evidentiary requirements. The example addresses the determination and proof of economic neediness under §53 Nr. 2 AO. The answer demonstrates multi-step reasoning, incorporation of administrative guidance (AEAO), numerical thresholds, and exceptions to documentation requirements.

**Accounting and booking questions**

This category targets the intersection of tax law and accounting practice. The example concerns private use of a company vehicle by a business owner and contrasts the 1% rule with the logbook method. The generated answer includes concrete booking entries, account references, and VAT treatment, reflecting realistic accounting workflows.

**Diversity-oriented questions**

Diversity questions are designed to expand topical and doctrinal coverage, including reorganization tax law and less frequently examined provisions. The example discusses §20 Abs. 2 Satz 5 UmwStG and the prevention of negative acquisition costs in reorganizations. The answer combines conceptual explanation with a numerical illustration to clarify the statutory mechanism.

Across all categories, the examples demonstrate that the Water Fountain Algorithm produces legally grounded, context-aware, and structurally diverse question-answer pairs. Generated answers consistently integrate statutory citations, case law references, numerical reasoning, and structured legal argumentation, closely resembling authentic examination and professional advisory material in German tax law.

# Supplementary Note 4: Water Fountain algorithm for synthetic German tax law data generation

Starting from a curated seed set of approximately 1,800 authentic German tax law examination questions, the algorithm iteratively retrieves authoritative legal sources via SearXNG, constructs bounded textual contexts using semantic similarity, generates grounded answers, filters insufficiently supported instances, and produces diversified follow-up questions under a fixed growth factor ($k = 3$). This process yields a large, validated dataset of legally grounded German tax law question-answer pairs for domain-specific language model training.

1. - Set iteration index $n \leftarrow 0$.
   - Set active question pool $Q_n \leftarrow Q_0$.
   - Initialize dataset $\mathcal{D} \leftarrow \emptyset$.

2. - While $n < n_{\max}$ and stopping criterion not met, do:
   (a) Retrieval and context construction: For each question $q \in Q_n$:
      i. Transform $q$ into a retrieval query and retrieve documents via SearXNG.
      ii. Segment retrieved documents into chunks $\{c_i\}$.
      iii. Compute semantic similarity:
      $$\text{sim}(c_i, q) = \frac{E(c_i) \cdot E(q)}{\| E(c_i) \| \| E(q) \|}.$$
      iv. Rank chunks by similarity and aggregate them sequentially until:
      $$\sum_{i=1}^{k} \text{tokens}(c_i) \leq N$$
      and adding the next chunk would exceed $N$.
      v. Denote the resulting bounded context as $C_q^*$.
   (b) Answer generation with sufficiency gating: Generate an answer: $a \leftarrow G(C_q^*, q)$.
      I. The generator is instructed to emit a predefined *INSUFFICIENT_CONTEXT* flag if the context does not support a valid answer.
      II. If the flag is present, discard $q$ and continue.
   (c) Quality and source validation:
      I. Count the number of independent retrieved sources supporting $C_q^*$.
      II. If this number is less than $S_{\min}$, discard $q$.
      III. Otherwise, accept the tuple $(q, a, C_q^*)$.
   (d) Dataset update:
      I. Assign a question type $t \in \mathcal{T}$ to $q$.
      II. Add $(q, a, C_q^*, t)$ to $\mathcal{D}$.
   (e) Controlled diversification ("fountain" step): For each accepted tuple, generate $k$ new questions:
      $$\{q_1', \ldots, q_k'\} \leftarrow G(C_q^*, \mathcal{T}),$$
      I. enforcing topical variation while remaining consistent with the legal context and taxonomy.
      II. Collect all generated questions into $Q_{n+1}$.
   (f) Iteration update:
      I. Set $Q_{n+1}$ as the active question pool and increment $n \leftarrow n + 1$.

3. - Stop when the target dataset size is reached, acceptance rate drops below a predefined threshold, or $n = n_{\max}$.

4. - Remove duplicates, overlaps with the original seed set, and any instances containing error flags or malformed outputs.

5. - Return the final validated dataset $\mathcal{D}$.

# Supplementary references

1. William G., C. *Sampling Techniques*. (Wiley, 1991).
2. Konietschke, F. & Pauly, M. Bootstrapping and permuting paired t-test type statistics. *Stat Comput* **24**, 283–296 (2014).
3. Dietterich, T. G. Approximate Statistical Tests for Comparing Supervised Classification Learning Algorithms. *Neural Computation* **10**, 1895–1923 (1998).
4. Benjamini, Y. & Hochberg, Y. Controlling the False Discovery Rate: A Practical and Powerful Approach to Multiple Testing. *Journal of the Royal Statistical Society Series B: Statistical Methodology* **57**, 289–300 (1995).
5. Bai, J. *et al.* Qwen technical report. *arXiv preprint arXiv:2309.16609* (2023).
6. Touvron, H. *et al.* LLaMA: Open and Efficient Foundation Language Models. Preprint at http://arxiv.org/abs/2302.13971 (2023).
7. Grattafiori, A. *et al.* The Llama 3 Herd of Models. Preprint at https://doi.org/10.48550/arXiv.2407.21783 (2024).
8. DeepSeek-AI *et al.* DeepSeek-R1: Incentivizing Reasoning Capability in LLMs via Reinforcement Learning. Preprint at https://doi.org/10.48550/arXiv.2501.12948 (2025).